\pdfoutput=1
\documentclass[sigconf]{acmart}

\AtBeginDocument{%
  }

\setcopyright{acmcopyright}
\copyrightyear{2023}
\acmYear{2023}
\setcopyright{rightsretained}
\acmConference[MM '23] {Proceedings of the 31st ACM International Conference on Multimedia}{October 29--November 3, 2023}{Ottawa, ON, Canada.}
\acmBooktitle{Proceedings of the 31st ACM International Conference on Multimedia (MM '23), October 29--November 3, 2023, Ottawa, ON, Canada}
\acmPrice{15.00}
\acmISBN{979-8-4007-0108-5/23/10}
\acmDOI{10.1145/3581783.3611907}
\usepackage{graphicx}
\usepackage{float}
\usepackage{subfig}

\usepackage{multirow}
\usepackage{makecell}

\begin{document}

% \title{Brighten-and-Colorize: Decoupled Personalized Enhancement for Low-Light Images}
\title{Brighten-and-Colorize: A Decoupled Network for Customized Low-Light Image Enhancement}

% \author{Anonymous submission\\
% paper ID xxxx}
\author{Chenxi Wang}
\affiliation{%
  \institution{Sun Yat-sen University}
  \city{Shenzhen}
  \state{Guangdong}
 \country{China}
}
\email{wangchx67@mail2.sysu.edu.cn}
\author{Zhi Jin}
\affiliation{%
  \institution{Sun Yat-sen University}
  \city{Shenzhen}
 \state{Guangdong}
   \country{China}
}
\email{jinzh26@mail2.sysu.edu.cn}
\authornote{Corresponding author.}
% \affiliation{%

% \renewcommand{\shortauthors}{Trovato \emph{et al.}}

\begin{abstract}
Low-Light Image Enhancement (LLIE) aims to improve the perceptual quality of an image captured in low-light conditions. Generally, a low-light image can be divided into lightness and chrominance components. Recent advances in this area mainly focus on the refinement of the lightness, while ignoring the role of chrominance. It easily leads to chromatic aberration and, to some extent, limits the diverse applications of chrominance in customized LLIE. In this work, a ``brighten-and-colorize'' network (called BCNet), which introduces image colorization to LLIE, is proposed to address the above issues. BCNet can accomplish LLIE with accurate color and simultaneously enables customized enhancement with varying saturations and color styles based on user preferences. Specifically, BCNet regards LLIE as a multi-task learning problem: brightening and colorization. The brightening sub-task aligns with other conventional LLIE methods to get a well-lit lightness. The colorization sub-task is accomplished by regarding the chrominance of the low-light image as color guidance like the user-guide image colorization. Upon completion of model training, the color guidance (i.e., input low-light chrominance) can be simply manipulated by users to acquire customized results. This customized process is optional and, due to its decoupled nature, does not compromise the structural and detailed information of lightness. Extensive experiments on the commonly used LLIE datasets show that the proposed method achieves both State-Of-The-Art (SOTA) performance and user-friendly customization. 
% \footnote{This work was supported by the National Natural Science Foundation of China under Grant No. 62071500. Supported by Sino-Germen Mobility Programme M-0421. }
\end{abstract}



\begin{teaserfigure}
  \includegraphics[width=\textwidth]{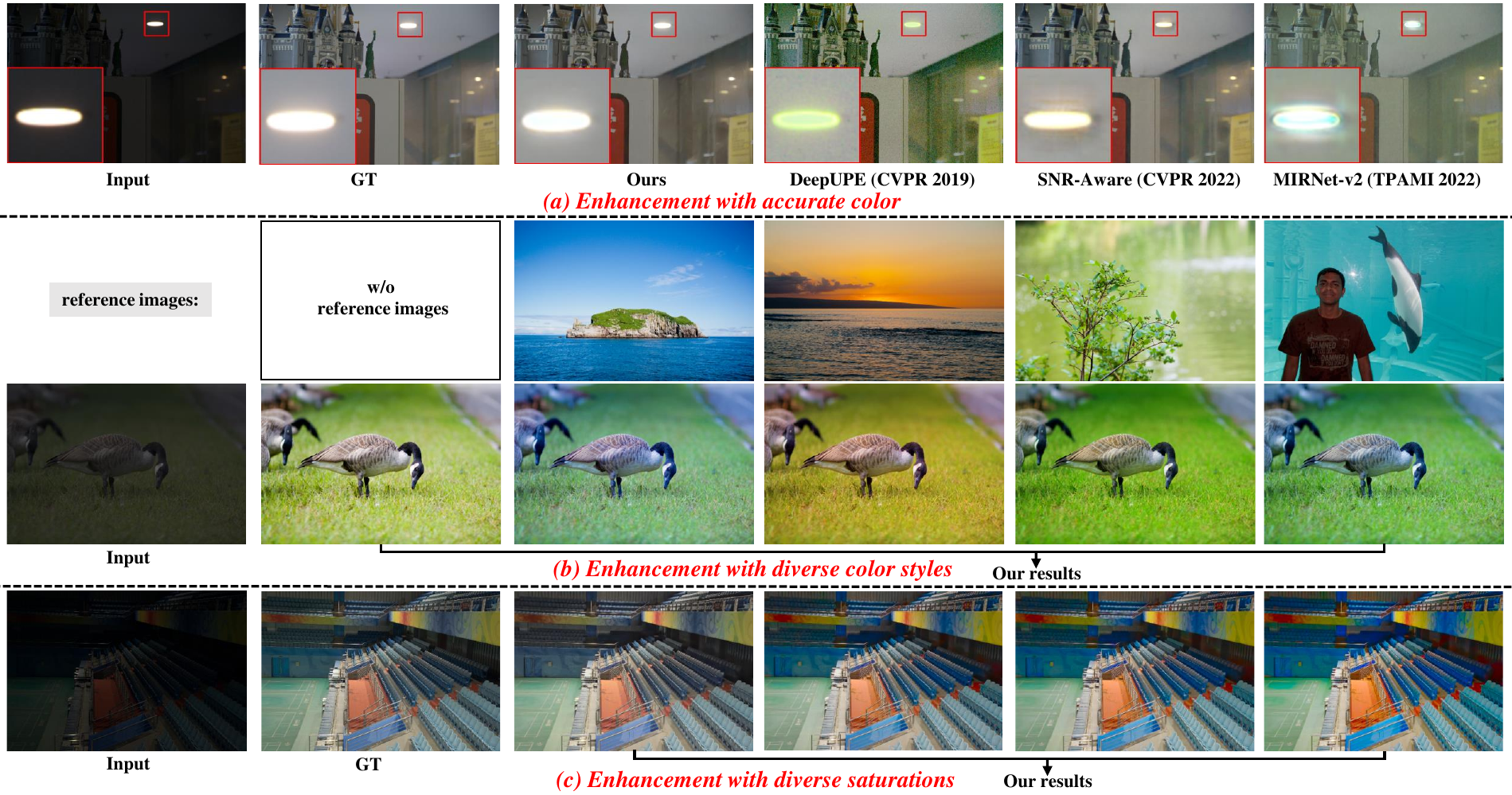}
  \vspace{-0.7cm}
  \caption{The enhanced results of the proposed method. (a) Compared with the results of current methods \cite{wang2019underexposed,xu2022snr,zamir2022learning}, which suffer from chromatic aberration, the proposed method reaches enhancement with accurate color. (b) The proposed method achieves customized enhancement based on reference images. (c) The proposed method can adjust the saturations of the enhanced results.}
  \label{fig:teaser}
\end{teaserfigure}

\maketitle
\section{Introduction}

Low-Light Image Enhancement (LLIE) is an important but challenging task in computer vision. It can not only improve the visual quality, but also be helpful for other high-level computer vision tasks (e.g., face detection \cite{yang2019joint}, action recognition \cite{pan2021view}, and object detection \cite{ren2015faster}). Until now, many traditional methods \cite{pizer1987adaptive,jobson1997multiscale,rahman2004retinex} and learning-based methods \cite{zhang2019kindling,lore2017llnet,xu2022snr,zamir2022learning, zamir2020learning, zhang2021beyond, wang2019underexposed, yang2021sparse, wei2018deep} have been proposed to enhance low-light images. However, most of them mainly focus on improving the quality of the lightness, which easily leads to serious loss of chrominance information and unpleasant visual results. Although some works try to apply additional constraints on chrominance information (e.g., the vectors of RGB channels \cite{wang2019underexposed}), they still hardly handle chrominance information well (as shown in Fig. \ref{fig:teaser} (a)). On the other hand, chrominance plays an important role in customized enhancement, however, recent customized methods \cite{zhang2021rellie,kim2020pienet,sun2021enhance,zheng2022enhancement,wang2022learning} do not notice this problem. iUP-Enhancer \cite{zheng2022enhancement} achieved chrominance customization by leveraging histograms in HSV color space, while the histograms as global information hardly guarantee local consistency. In summary, existing (customized) LLIE methods do not pay enough attention to the importance of chrominance information.

Meanwhile, image colorization, as an active research field that aims to predict lost chrominance information on the given lightness, has two key challenges: 1) \textit{how to ensure boundaries of the generated chrominance}, 2) \textit{how to solve the problem that one object can be filled by multiple colors (e.g., a balloon can be colorized red or blue)}. Some image colorization methods tackle these problems by introducing semantic information \cite{zhao2020pixelated,su2020instance}, which can provide the boundaries or object information, and the user interactions (e.g. color strokes \cite{zhang2017real} or reference image \cite{he2018deep,lu2020gray2colornet}), which are helpful for generating the certain colors. However, we observe that although the chrominance of low-light image is unsaturated, it contains both boundary information and some color hints. 

Hence, we novelly propose to regard the chrominance prediction in LLIE as a colorization problem, which takes the low-light chrominance as a guidance to recover proper colors. In this way, LLIE is decoupled into the brightening and colorization sub-tasks. Since the input of the colorization process is usually carried out on grey images under normal light, an intuitive implementation is to first brighten the low-light lightness and then colorize it. However, this two-step approach is tedious and lacks information interactions between two sub-tasks. In this work, a new ``brighten-and-colorize'' paradigm, called BCNet is proposed. BCNet adopts multi-task learning architecture, which contains one encoder and two task-specific decoders. The encoder takes the low-light lightness as the input, while the two task-specific decoders aim to output predicted lightness and chrominance, respectively. By training two sub-tasks simultaneously, the brightening sub-task can provide information of normal-light lightness to the colorization sub-task. Besides, the color classification loss \cite{zhang2016colorful}, which transforms colorization to a classification task, is introduced to relieve color vanishment in the colorization sub-task.

Since the chrominance of the low-light image serves as color guidance for colorization, users can modify it to generate customized results in the testing phase. In our work, we introduce two customized operations. Firstly, users can change the color style of the guidance based on a reference image, resulting in diverse color styles (see Fig. \ref{fig:teaser} (b)). Secondly, users can adjust the saturation of the guidance coarsely to achieve the enhancement with different saturations (see Fig. \ref{fig:teaser} (c)). Note that the key process of customized enhancement in this work is to generate customized color guidance for the colorization sub-task, some simple non-learning operators are sufficient to produce pleasing results. Meanwhile, benefiting from the decoupled nature, the details of the lightness component can be preserved well during any customized processes.

To sum up, our contributions are four folds: \textbf{1) We decouple LLIE into brightening and colorization sub-tasks by introducing image colorization. 2) A multi-task learning architecture called BCNet is proposed to implement this decoupled enhancement. 3) Based on the property of the colorization sub-task, we provide a new solution for customized LLIE. 4) Extensive experiments on popular LLIE datasets demonstrate the proposed method reaches SOTA
performance and performs well on customized LLIE.}

\begin{figure*}
  \centering
  \captionsetup[subfloat]{labelformat=empty}
  
  \subfloat[]{
    %\hspace{0.2in}
    \includegraphics[width=1.\textwidth]{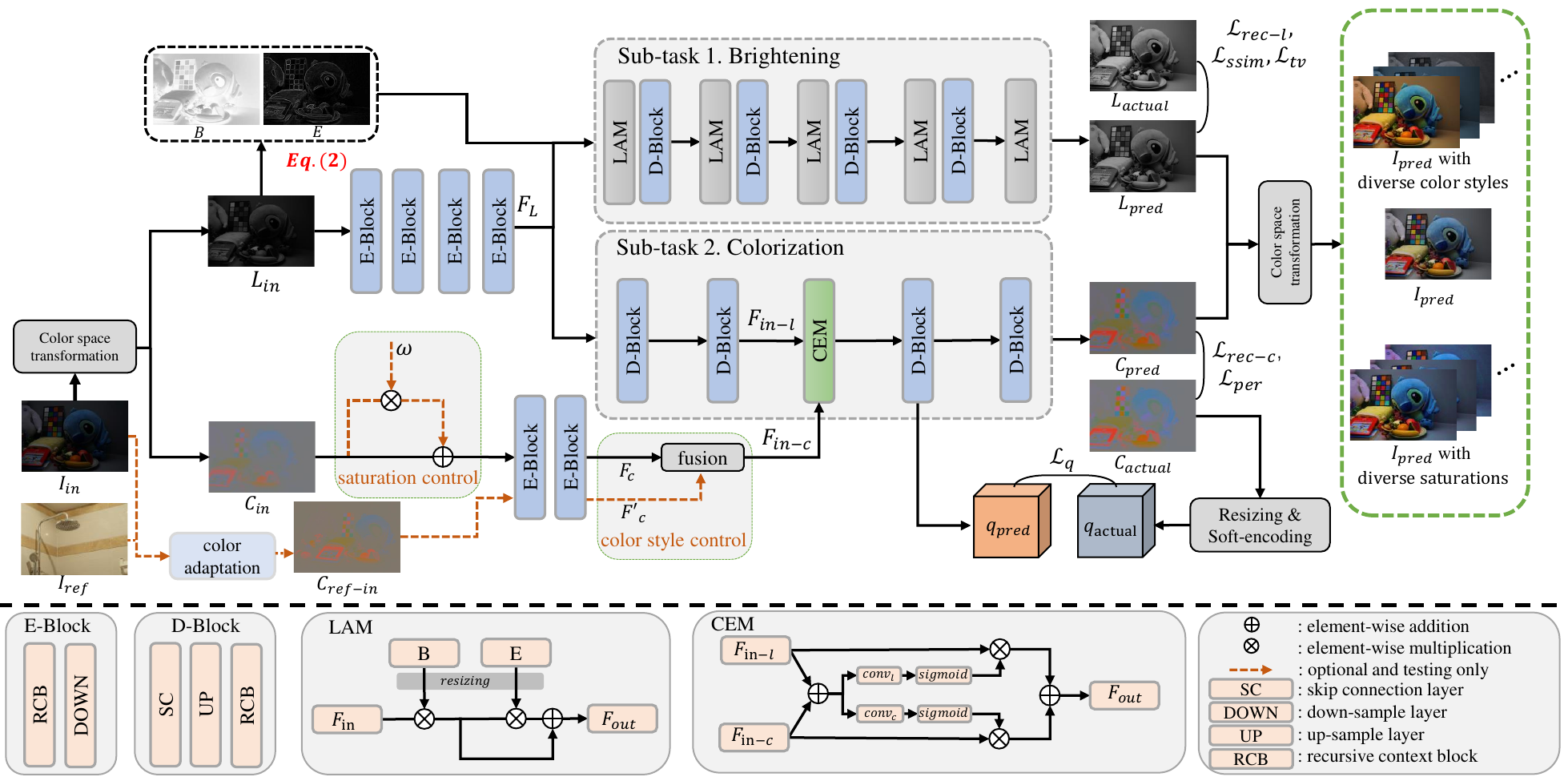}
  }

 \vspace{-0.7cm}
  \caption{Framework of the proposed method. The proposed method adopts multi-task learning architecture, containing one encoder and two task-specific decoders. It implements two sub-tasks: brightening and colorization. After model training is completed, the input chrominance can be manipulated by users to achieve enhancement with diver styles and saturations.}
    \label{fig:pipeline}
    \vspace{-0.3cm}
\end{figure*}

\section{Related Work}
\subsection{Low-Light Image Enhancement}
% \noindent\textbf{ }

Traditional low-light image enhancement (LLIE) methods can be broadly categorized into two types: Histogram Equalization (HE)-based methods \cite{pizer1987adaptive} and Retinex-based methods \cite{jobson1997multiscale,rahman2004retinex}. HE-based methods aim to flatten the histograms of low-light images, while Retinex-based methods attempt to decompose low-light images into illumination and reflectance components. However, these traditional methods often struggle with noise and color. In recent years, Deep Neural Networks (DNNs) have shown great potential for image enhancement \cite{lore2017llnet,zhang2022deep,huang2022deep,li2023embedding,kim2021representative,jin2020dual,jin2019flexible,kim2020global,wu2022lightweight}. The first Convolutional Neural Network (CNN) architecture designed for LLIE was proposed by Lore \emph{et al.} \cite{lore2017llnet}, and since then, several other CNN-based architectures \cite{gharbi2017deep,kim2021representative,kim2020global,wu2022lightweight} have been proposed. To achieve more realistic results, Retinex theory has been incorporated into DNNs \cite{wei2018deep,zhang2019kindling,zhang2021beyond,yang2021sparse}. Methods in \cite{guo2020zero,wang2019underexposed,zhang2022deep} apply an additional constraint on color information. Besides, many unsupervised methods \cite{jiang2021enlightengan,guo2020zero,liu2021retinex,ma2022toward} have been proposed to eliminate the requirement of paired training data. However, these methods often do not make optimal use of chrominance. 

Another line of LLIE methods aims to achieve customized enhancement to satisfy the different preferences of users. The first method to introduce customization to LLIE was PieNet \cite{kim2020pienet}, which achieved customized enhancement by extracting style vectors from reference images. TSFlow \cite{wang2022learning} introduced style vectors to normalizing flow \cite{kingma2018glow} to obtain diverse results. However, these "black-box" processes may lead to poor flexibility for user customization. ReLLIE \cite{zhang2021rellie} achieved enhancements with different brightness by introducing deep reinforcement learning, but it ignores chrominance information. iUP-Enhancer \cite{zheng2022enhancement} provided a "white-box" approach to accomplish customized enhancement by leveraging histogram information in the HSV color space. While histograms as a form of global information may hardly ensure local consistency. Compared to existing customized LLIE methods, BCNet provides a new perspective by using image colorization to achieve customization. This approach preserves lightness details well and allows for accurate chrominance.
\vspace{-0.2cm}
\subsection{Image Colorization}
% \noindent\textbf{}

Image colorization techniques can be classified into two categories: automatic colorization and user-guided colorization. Automatic colorization \cite{zhang2016colorful,guadarrama2017pixcolor,kumar2021colorization} aims to colorize grayscale images without any external color guidance. Zhang \emph{et al.} \cite{zhang2016colorful} transformed image colorization into a classification task to generate diverse colorization results. Kumar \emph{et al.} \cite{kumar2021colorization} introduced transformer \cite{vaswani2017attention} to image colorization. However, automatic colorization methods often suffer from color ambiguity. To address this issue, user-guided image colorization methods were proposed. Zhang \emph{et al.} \cite{zhang2017real} utilized color strokes as color guidance, while He \emph{et al.} \cite{he2018deep}, Lu \emph{et al.} \cite{lu2020gray2colornet}, and Yin \emph{et al.} \cite{yin2021yes} used exemplar images as color references. Moreover, to achieve more precise colorization, some methods \cite{su2020instance,zhao2020pixelated} incorporated semantic information to provide object boundaries and object-specific colorization. In this work, we present a user-guided colorization method that leverages the chrominance information of low-light images to provide not only color guidance but also object boundary information. We also introduce the concept of color classification \cite{zhang2016colorful} and apply both regression and classification constraints to achieve more robust colorization.
\section{Method}
\subsection{``Brighten-and-Colorize''}

To accomplish the ``brighten-and-colorize'' enhancement, we first decompose the image into lightness and chrominance. In widely used RGB color space, every single channel contains part of lightness and chrominance, which means they are inseparable in the RGB color space. Therefore, to separate lightness and chrominance, we transform the image from RGB color space to CIELAB color space, where the ``L'' channel represents lightness and the ``AB'' channels represent chrominance. Note that we follow the commonly used color space transform function in the image colorization field.

After decomposing the image, the LLIE task is decoupled into two sub-tasks: brightening and colorization. The brightening itself can be regarded as an LLIE problem that predicts lightness with well exposure and details. The colorization aims to predict realistic and accurate chrominance based on the normal-light lightness information and low-light chrominance like user-guide image colorization methods.

\subsection{Network Architecture}
In this sub-section, we present the details of the proposed method BCNet shown in Fig. \ref{fig:pipeline}. BCNet adopts multi-task learning architecture and contains an encoder and two task-specific decoders. Given a low-light image $I_{in}\in \mathbb{R}^{H\times W\times3 }$, we first decompose it into a lightness map $L_{in}\in \mathbb{R}^{H\times W\times1 }$ and a chrominance map $C_{in}\in \mathbb{R}^{H\times W\times2 }$. Then, the encoder takes $L_{in}$ as input and two task-specific decoders output a predicted lightness $L_{pred}\in \mathbb{R}^{H\times W\times1 }$ and a predicted chrominance $C_{pred}\in \mathbb{R}^{H\times W\times2 }$, respectively. The constraints are applied with lightness of ground-truth $L_{actual}\in \mathbb{R}^{H\times W\times1 }$ and chrominance of ground-truth $C_{actual}\in \mathbb{R}^{H\times W\times2 }$. In the following parts, we illustrate the reasonability for adopting the multi-task learning design and the details of two task-specific decoders.

\subsubsection{Multi-Task Encoder}

The structure content of the lightness is essential for the brightening and colorization sub-tasks. It determines the network how to brighten and where to colorize. In this work, the brightening sub-task aims to recover the clear structure content of lightness, and the colorization sub-task predicts the lost chrominance based on the clear structure content. Due to different inputs, the two sub-tasks are not in the typical design of multi-task learning. However, when training the brightening sub-task, the encoder can learn the information of the clear structure content, which is necessary for the colorization sub-task. By sharing the same encoder, this necessary information for the colorization sub-task can be obtained from the brightening sub-task. Hence, it still can be regarded as a multi-task learning problem. Further, this design not only encourages the information interactions between two sub-tasks but also makes the network more efficient. 

Referring to Fig. \ref{fig:pipeline}, the multi-task encoder takes the low-light lightness image as input and contains four encoding blocks (E-Block). Every encoding block consists of a Recursive Context Block (RCB) \cite{zamir2022learning}, which provides a more effective feature extraction, and a down-sample layer. In this way, we can extract the features $F_{L}$ from $L_{in}$ as:
\begin{equation}
    F_{L}=Encoder(L_{in})
\end{equation}

Note that the output of every encoding block is transmitted to the corresponding decoding block in two sub-tasks by skip-connection layers.

\vspace{-0.2cm}
\subsubsection{Lightness Decoder}
The lightness decoder is adopted in the brightening sub-task. It contains four decoding blocks (D-Block) and five Lightness Adjustment Modules (LAMs). Every decoding block consists of a skip-connection layer, an up-sample layer, and an RCB. The LAM is proposed for guiding the network to improve contrast and preserve better details. Empirically, the darker regions of a low-light image need to be brightened more than brighter regions, and the edges information is essential for an enhanced image. Hence, LAM first takes the inverted map $B$ of $L_{in}$ as the attention prior to indicate where needs more contrast improvement. Then, it utilizes the edge map $E$ of $L_{in}$ to help preserve edges. As shown in Fig. \ref{fig:pipeline}, the LAM can be formulated as:
\begin{equation}
    B=1-L_{in},E=edge(L_{in})
\end{equation}
\begin{equation}
   F_{out}=F_{in}\times \psi (B) \times(1+\psi(E))
\end{equation}
where $F_{in}$ and $F_{out}$ are input and output features, respectively. $\psi(.)$ represents resizing operation (bilinear is adopted in this work). $edge$ denotes an edge extractor, which is a $sobel$ operator with kernel size of 3. Then, the predicted lightness $L_{pred}$ can be obtained by:
\begin{equation}
   L_{pred} = Brightening(F_{L}, B, E)
\end{equation}

\begin{figure}
    \centering
    \includegraphics[width=0.9\columnwidth]{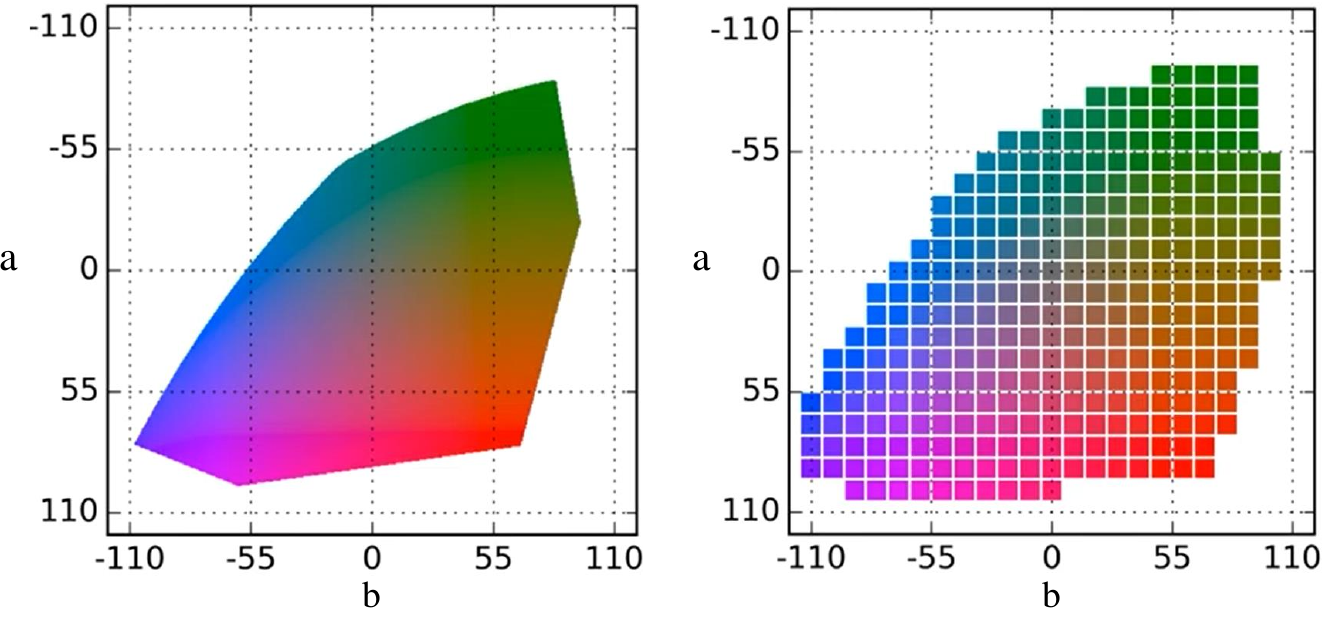}
     \vspace{-0.3cm}
    \caption{The quantization operation \cite{zhang2016colorful} in CIELAB color space. The continuous colors (left) are quantized to 313 discrete colors (right) with a grid size of 10. 
    % The image is got from http://richzhang.github.io/colorization/.
    }
    \label{fig:ab_gamut}
    \vspace{-0.3cm}
\end{figure}

\begin{figure*}
    \centering
    \includegraphics[width=1.\textwidth]{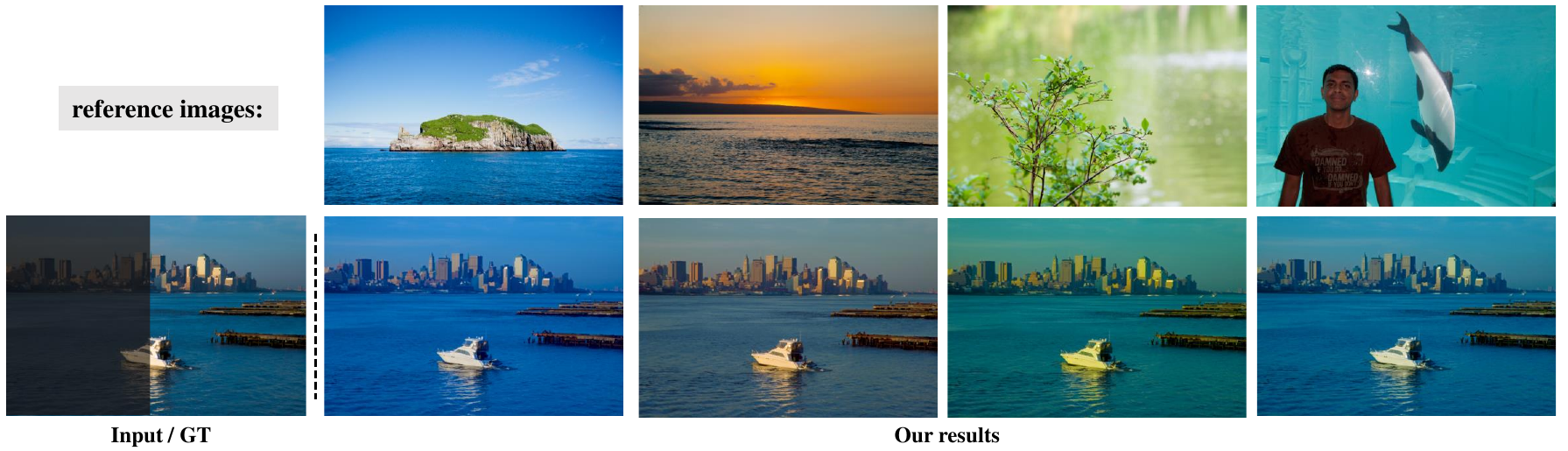}
     \vspace{-0.3cm}
    \caption{The visual results of color style control. It can be seen that the enhanced results have different color styles based on different reference images.}
    \label{fig:diver_style}
    \vspace{-0.2cm}
\end{figure*}
\begin{figure*}
    \centering
    \includegraphics[width=1.\textwidth]{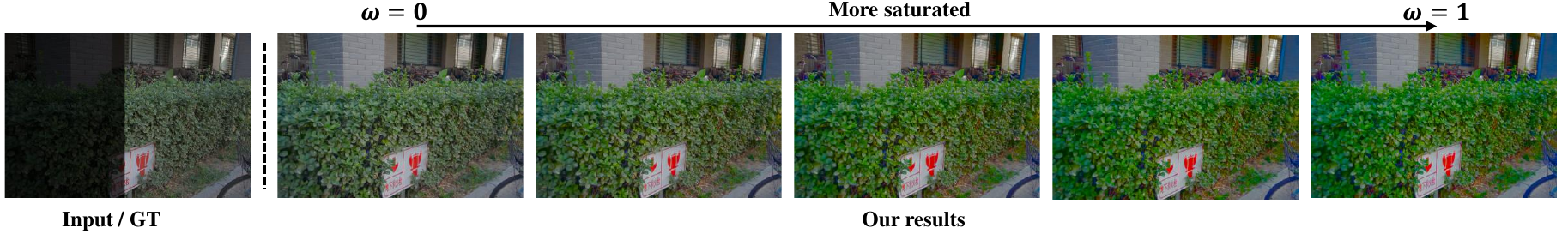}
     \vspace{-0.3cm}
    \caption{The visual results of saturation control. It can be seen that the result becomes more saturated as the value of $\omega$ goes up.}
    \label{fig:colorful}
    \vspace{-0.2cm}
\end{figure*}

\subsubsection{Chrominance Decoder}

The chrominance decoder is adopted in the colorization sub-task, which contains four decoding blocks and a Color Embedding Module (CEM).
\vspace{0.1cm}

\noindent\textbf{Colorization by the low-light chrominance.} As mentioned before, colorization is an ill-posed problem, since the color of a certain object is ambiguous. Existing image colorization methods address this issue through the user-guide strategy that introduces an exemplar image or color strokes as color hints. The exemplar image can offer a specific color style, and the color strokes provide approximate guidance. Besides, the colorization methods easily colorize one object beyond its boundary, and this problem can be relieved by introducing semantic information, which reflects the boundaries of objects, to the colorization process. However, different from the original image colorization tasks, in this work, the low-light images already contain part of the chrominance information. Although it is unsaturated, it can provide enough clues to help generate the proper chrominance. For example, the chrominance $C_{in}$ in Fig. \ref{fig:pipeline} is from the input low-light image, it still reflects the approximate color tone and shapes of the ground-truth chrominance $C_{actual}$. 

Therefore, we utilize this information by embedding the features of $C_{in}$ to the chrominance decoder. Inspired by \cite{li2019selective}, we regard features extracted from lightness and chrominance as multi-scale (i.e., multiple information sources) features and propose a CEM to accomplish feature fusion. As shown in Fig. \ref{fig:pipeline}, given the input features $F_{in-l}$ extracted from $L_{in}$ and $F_{in-c}$ extracted from $C_{in}$, CEM first get the summation feature $\hat{F_{in}}$ of two features by element-wise addition as:
\begin{equation}
   \hat{F_{in}}  = F_{in-l}+F_{in-c}
\end{equation}

Then, we calculate affinity matrices $A_{l}, A_{c}$, which are used to embed chrominance features to the colorization process, by two convolutional layers and $sigmoid$ activations as:
\begin{equation}
   A_{l} = sigmoid(conv_{l}(\hat{F_{in}})),A_{c} = sigmoid(conv_{c}(\hat{F_{in}}))
\end{equation}

Finally, the output feature $F_{out}$ of CEM is obtain by:
\begin{equation}
   F_{out} = F_{in-l} \times A_{l} + F_{in-c} \times A_{c}
\end{equation}

% \vspace{0.1cm}
\noindent\textbf{Colorization free from color vanishment.} Regarding colorization as a regression task easily results in color vanishment. For example, the day sky is blue and the dusk sky is red, while the mean of red and blue is gray. To relieve this problem, we quantize entire colors in CIELAB color space to 313 categories with a grid size of 10 (as shown in Fig. \ref{fig:ab_gamut}) and calculate the classification loss followed by \cite{zhang2016colorful}. It is an effective solution for color vanishment. Note that the classification output $q_{pred}\in \mathbb{R}^{h\times w\times 313 }$ hardly reaches the original size of the image limited by memory. We set $h$ and $w$ to a quarter of the original size. However, it may affect the enhanced result at the pixel-level. On the other hand, the quantized operation can also lead to performance degradation. Therefore, we apply constraints on color classification output and chrominance map simultaneously, which is helpful for generating various colors and more precise colors at pixel-level, respectively. The predicted chrominance map $C_{pred}$ and color classification output $q_{pred}$ can be expressed by:
\begin{equation}
   C_{pred},q_{pred}=Colorization(F_{L},C_{in})
\end{equation}

\subsection{Loss Functions}
We apply different constraints in the brightening and colorization sub-tasks based on their different characteristics. For lightness losses, Charbonnier loss \cite{lai2018fast} is first applied as the reconstruct loss $\mathcal{L}_{rec-l}$ to supervise lightness reconstruction at pixel-level, which can be defined as:
\begin{equation}
    \mathcal{L}_{rec-l}= \sqrt{\left \| L_{actual}-L_{pred} \right \| _{2}+\epsilon ^{2}  } 
\end{equation}
where $\epsilon$ is set to $10^{-3}$ empirically. Then, the SSIM \cite{wang2004image} loss $\mathcal{L}_{ssim}$ and total variation \cite{chambolle2004algorithm} loss $\mathcal{L}_{tv}$ are applied to get a better structural details. For chrominance losses, 
as analyzed in \cite{zhang2017real,zhang2016colorful}, L2 loss is not robust for colorization due to the inherent multi-modal nature of colorization. In this work, we apply L1 loss as chrominance reconstruction loss $\mathcal{L}_{rec-c}$, which is described as:
\begin{equation}
    \mathcal{L}_{rec-c}=\left \| C_{actual}-C_{pred} \right \| _{1} 
\end{equation}

Next, the perceptual loss $\mathcal{L}_{per}$, which constrains in features extracted from VGGNet\cite{simonyan2014very}, is adopted to get better visual results. 

Then, color classification loss $\mathcal{L}_{q}$ is applied to get various colors. The $\mathcal{L}_{q}$ is defined as:
\begin{equation}
    \mathcal{L}_{q}=\mathcal{H}(q_{actual},q_{pred})
\end{equation}
where $\mathcal{H}(.)$ is a 2-D cross-entropy loss function. $q_{actual}$ is obtained by soft-encoding resized $C_{actual}$ as \cite{zhang2016colorful}.  

The total loss of BCNet $\mathcal{L}_{total}$ can be formulated as:
\begin{equation}
\begin{split}
    \mathcal{L}_{total}=&\lambda _1\mathcal{L}_{rec-l}+\lambda _2\mathcal{L}_{ssim}+\lambda _3\mathcal{L}_{tv}+ \\ 
    &\lambda _4\mathcal{L}_{rec-c}+\lambda _5\mathcal{L}_{per}+\lambda _6\mathcal{L}_{q}
\end{split}
\end{equation}
where $\lambda _1,\lambda _2,\lambda _3,\lambda _4,\lambda _5,\lambda _6$ are weigh factors and  set to 1, 0.1, 0.01, 1, 0.01, 0.01, respectively.

\subsection{Customized Enhancement}
The chrominance of low-light images is served as color guidance to help the colorization sub-task. It provides two essential pieces of information: 1) boundaries information for guiding where to colorize, and 2) color hints for guiding what color to be used. When model training is completed, users can modify the color guidance to achieve customized enhancement. This process only needs to maintain the boundaries information of color guidance to ensure the position of predicted chrominance. Meanwhile, the lightness component is lossless after customized operations since the guidance only affects the colorization sub-task. In this work, two customized operations, which achieve color style control (see Fig. \ref{fig:diver_style}) and color saturation control (see Fig. \ref{fig:colorful}), are proposed as follows.

\subsubsection{Color Style Control}
\begin{figure}
    \centering
    \includegraphics[width=0.9\columnwidth]{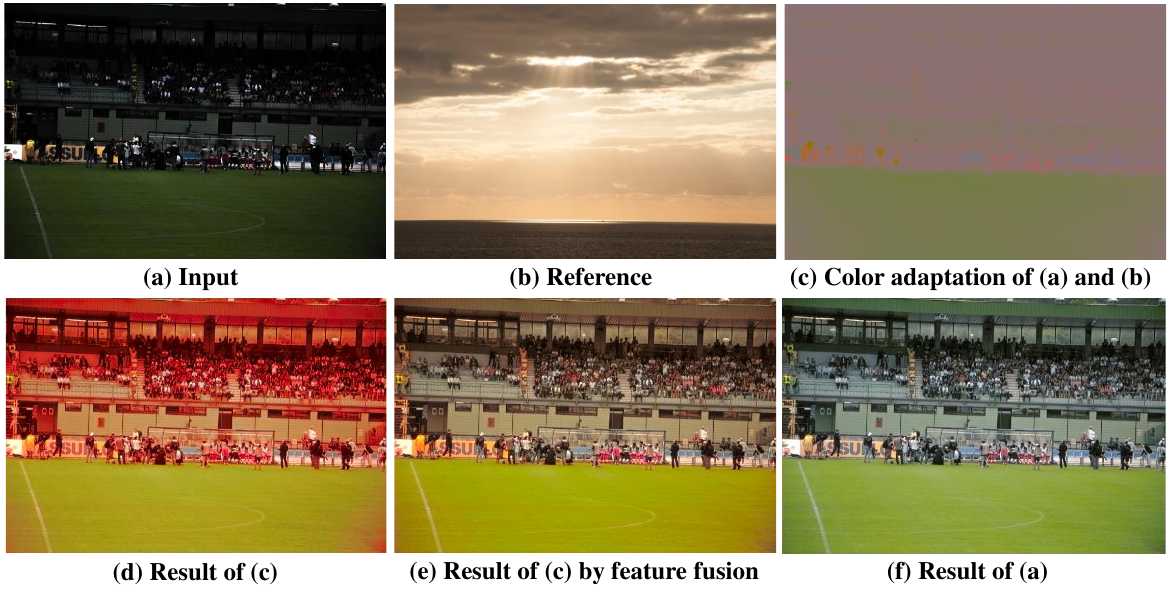}
     % \vspace{-0.3cm}
    \caption{ The process of color style control. The input image (a) and reference image (b) first through a color adaptation module \cite{reinhard2001color} to get customized color guidance (c). (d) is the result of directly embedding (c) to the colorization process, (e) is the result of embedding (c) to the colorization process with a feature fusion module, and (f) is the result of directly embedding the original chrominance of (a) to the colorization process. By comparing (d), (e), and (f), we can find the color style of (b) is transferred to (e) and the feature fusion module is helpful for keeping the color details.}
    \label{fig:diver_style2}
    % \vspace{-0.3cm}
\end{figure}

To generate results with diverse color styles, the color hints provided by guidance can be modified. However, since the boundary information needs to be preserved, manually modifying them may be complicated. One effective solution is to introduce a reference image. There are many color style transfer works \cite{he2018deep,lu2020gray2colornet,yin2021yes} that can transmit the color style from a reference image to the input image. In our case, since the modification is applied to the guidance, we do not need to explore a complex and accurate transfer. Therefore, we adopt a traditional non-learning color transfer method \cite{reinhard2001color}, which achieves color transfer by transforming images to an orthogonal color space (the detailed implementation can be found in supplementary materials), to obtain customized color guidance (as shown in Fig. \ref{fig:diver_style2} (c)). However, this traditional method may hardly perform well in some color details when the input and reference images are irrelevant (as shown in Fig. \ref{fig:diver_style2} (d)). To address this issue, we fuse the features $F_{c}$ extracted from $C_{in}$ into the features $F'_{c}$ extracted from $C_{ref-in}$ to retain the color details (as shown in Fig. \ref{fig:diver_style2} (e)). The fusion process is defined as $(1-\gamma) \times F_{c} + \gamma \times F'_{c}$, where $\gamma$ is a hyper-parameter that balances the two features and is set to 0.7 for testing and 0 for training. 

\subsubsection{Saturation Control}
The color guidance provides what color should be used, besides, its intensity also can affect the saturation of output. To preserve the boundaries information, a simple amplified operation is designed as shown in ``saturation control'' of Fig. \ref{fig:pipeline}. The amplified color guidance is obtained by $C_{in}\times (1+\omega)$, where $\omega$ is a hyper-parameter to control the saturation. The bigger $\omega$ represents the result is more saturated and colorful (see Fig. \ref{fig:colorful}). Actually, $\omega$ can be any value, we empirically set $\omega \in [0,1]$. When $\omega < 0 $ or $\omega > 1 $, the results may be over dull or saturated.

Note that the above customized operations are not involved in model training. They are optional and only run in the testing phase.

\section{Experiment}

\begin{table*}
\caption{Quantitative comparison on the LOL-real \cite{yang2021sparse} and FiveK \cite{bychkovsky2011learning}.
The best results are boldfaced and the second-best ones are
underlined.}
 \vspace{-0.25cm}
    \label{tab:experiments}
       \resizebox{\textwidth}{!}{
\renewcommand{\arraystretch}{1} % 可以调节, 1.2指高度是默认的1.2倍
    \centering
      \begin{tabular}{c c c c c c c c c c c c}
         \toprule

     \multirow{2}{*}{Methods} &\multicolumn{5}{c}{LOL-real \cite{yang2021sparse}}&\multicolumn{5}{c}{FiveK \cite{bychkovsky2011learning}}&\multirow{2}{*}{Size (M)} \\
      \cmidrule(r){2-6} \cmidrule(r){7-11} 
     &   PSNR $\uparrow$    &   SSIM $\uparrow$ & $\Delta E\_{ab} $$\downarrow $     &   LPIPS $\downarrow $ &   CSE (ratio) $\downarrow$ &   PSNR $\uparrow$    &   SSIM $\uparrow$ &$ \Delta E\_{ab}$ $\downarrow $     &   LPIPS $\downarrow $ &   CSE (ratio) $\downarrow$ &   \\
      % &   PSNR    &   SSIM  & \Delta E_{ab}     &   LPIPS &   CSE (ratio)  &   PSNR     &   SSIM  & \Delta E_{ab}    &   LPIPS  &   CSE (ratio)  &   \\
      % \hline
        \midrule
      %\cmidrule(5-8)
      DRD\cite{wei2018deep} & 16.08 & 0.6555 & 22.35 & 0.2364 & 2.53 & 21.68 & 0.8604 & 10.52 & 0.0574 & 2.18 & \textbf{0.86}\\
      Kind\cite{zhang2019kindling} & 20.01 & 0.8412 & 12.53 & 0.0813& 1.30 & 20.71 & 0.8835 & 10.75 & 0.0480 & 1.38 & 8.02\\
      Kind++\cite{zhang2021beyond} & 20.59 & 0.8294  & 12.51 & 0.0875 & 1.18 & 19.71 & 0.8640 & 14.05 & 0.0574 & 2.12 & 8.27\\
      MIRNet\cite{zamir2020learning} & \underline{22.11} & 0.7942 & \underline{10.11} & 0.1448& 1.35 & 24.41 & 0.9097 & 7.90 &  0.0344& \underline{1.33} & 31.79\\
      EnGAN\cite{jiang2021enlightengan} & 18.64 & 0.6767 & 17.73  & 0.1512& 6.18 & 15.38 & 0.7752 & 18.89 & 0.0984 & 2.97 & 8.64\\
      DeepUPE\cite{wang2019underexposed} & 18.68 & 0.5791 & 17.54 & 0.1868 & 9.48 & 24.24 & 0.8957 & 8.16 & 0.0440& 2.15 & \underline{0.99} \\
     % Zero-DCE\cite{guo2020zero} & 18.41 & 0.5677 & 17.20 & 0.2270 & 13.75 & 14.67 & 0.7448 & 22.0155 & 0.1491 & 8.92 & \textbf{0.0056}\\
      DeepLPF\cite{moran2020deeplpf} & 20.03 & 0.7819 & 12.58 & 0.1460 & 3.23 & 24.74 & 0.9170  & 7.50 & 0.0570& 1.44  & 1.72 \\
      UEGAN\cite{ni2020towards} & 20.30 & 0.7417 & 14.69  & 0.1464 & 12.97  & 23.00 &  0.8717 & 9.96 & 0.0503& 3.84 & 4.16 \\
      SGM\cite{yang2021sparse} & 20.06 & 0.8158 & 11.36 & 0.0727 & 1.33 & 22.57 & 0.8823& 9.36 & 0.0557 & 4.23  & 2.31 \\
    %   DRBN & 20.11 & 0.8534 & 59.42 & 1470.91 & 0.0611\\
      MIRNet-v2\cite{zamir2022learning} & 21.83 & 0.8455 & 11.47 & \underline{0.0666} & 1.49 & 25.04 & 0.9188 & 8.05 & 0.0357& 1.54 & 5.86\\
      SNR-Aware\cite{xu2022snr} & 21.48 & \underline{0.8478} & 10.58 &  0.0740& \underline{1.14} & \underline{25.41} & \underline{0.9234} & \underline{7.24} & \underline{0.0293} & 1.49 & 39.12\\
     \midrule
     \multicolumn{1}{c}{Ours} & \textbf{23.27} & \textbf{0.8637} & \textbf{8.97} & \textbf{0.0566}& \textbf{1.00} & \textbf{25.74} & \textbf{0.9285} & \textbf{6.77} & \textbf{0.0291}& \textbf{1.00}  & 6.84\\
     \bottomrule
     
    \end{tabular}}
\vspace{-0.3cm}
\end{table*}

\subsection{Datasets and Implementation Details}
LOL-real \cite{yang2021sparse} and MIT-Adobe-FiveK \cite{bychkovsky2011learning} (for short called FiveK) are adopted as our experimental datasets. LOL-real is captured in the real world by changing exposure time and ISO. It contains 689 pairs of low-/normal-light images for training and 100 pairs for testing. FiveK contains 5,000 raw images and corresponding five high-quality versions retouched by experts. The version retouched by expert C is adopted as the ground-truth and it is divided into 4,500 training pairs and 500 testing pairs following \cite{wang2019underexposed,zamir2022learning}. 

BCNet is implemented in PyTorch and trained in an RTX2080Ti GPU with batch size of 8. The learning rate is initiated to $2.0\times10^{-4}$ and a multi-step scheduler is adopted. Adam \cite{kingma2014adam} with momentum of 0.9 is adopted as the optimizer. Input training images are randomly cropped to $256 \times 256$ and rotated by multiples of 90 degrees.

\subsection{Comparison with Recent LLIE Methods}

\begin{figure}
    \centering
    \includegraphics[width=0.9\columnwidth]{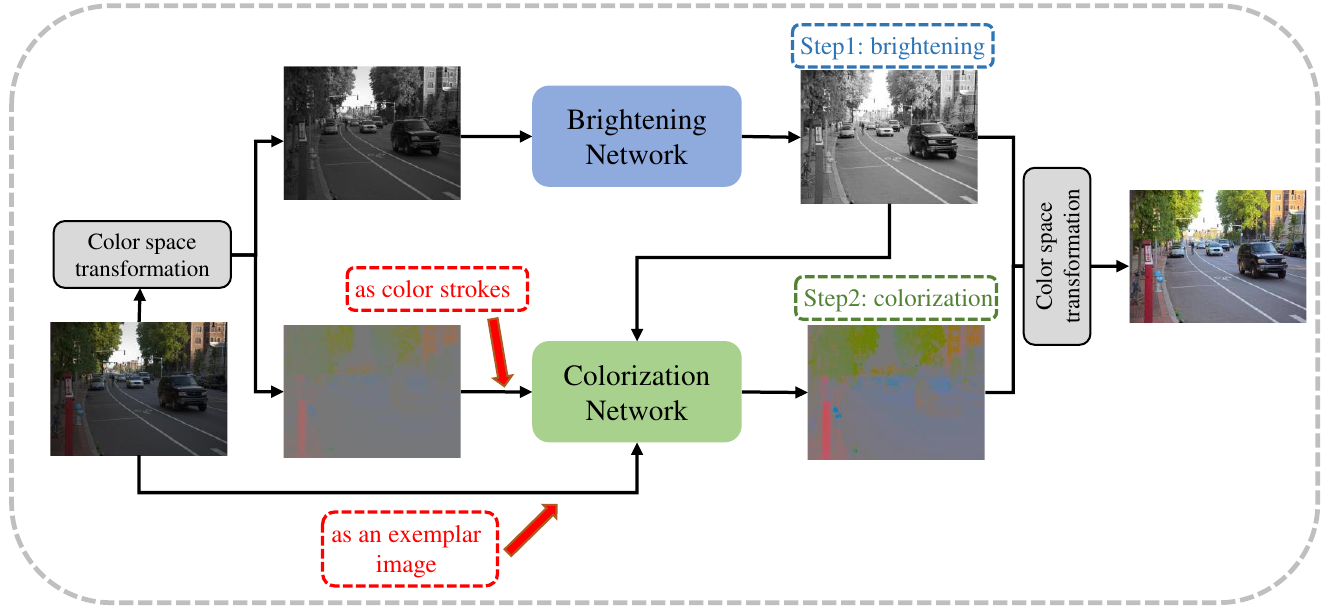}
     \vspace{-0.3cm}
    \caption{The illustration of the two-step ``brighten-and-colorize''. Note that the exemplar-based colorizer takes the low-light image as color hints and the stroke-based colorizer takes the chrominance of the low-light image as color hints.}
    \label{fig:b-and-c}
    \vspace{-0.5cm}
\end{figure}

\begin{figure*}
    \centering
    \includegraphics[width=1.\textwidth]{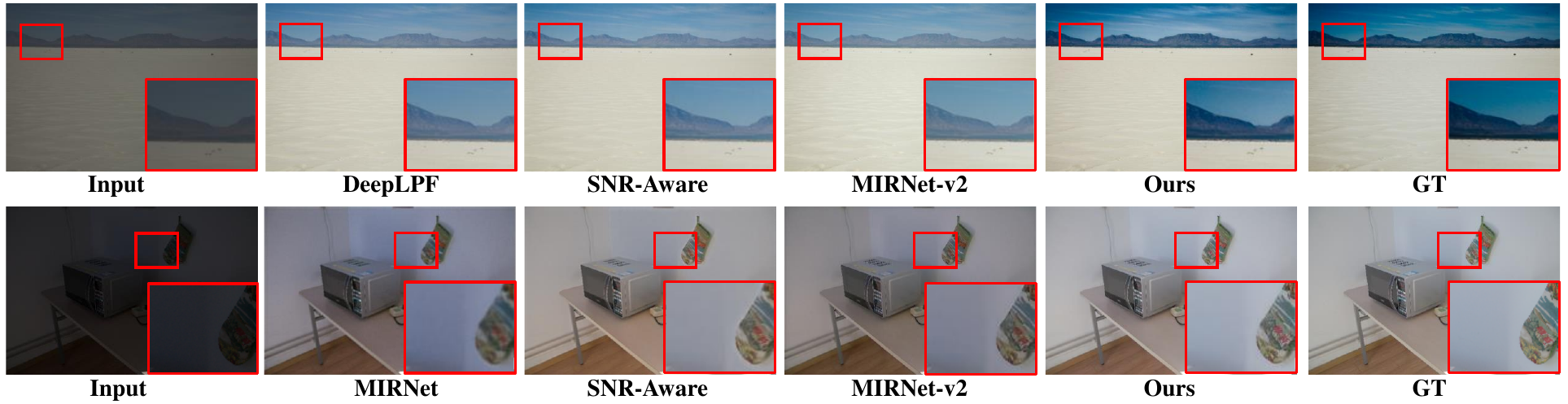}
     \vspace{-0.7cm}
    \caption{The qualitative comparison on FiveK \cite{bychkovsky2011learning} (the first row) and LOL-real \cite{yang2021sparse} (the second row). It can be seen that the proposed method reaches the best visual results.}
    \label{fig:5k_lol}
    \vspace{-0.5cm}
\end{figure*}

We compare BCNet with recent SOTA LLIE methods, which include DRD \cite{wei2018deep}, Kind \cite{zhang2019kindling}, Kind++ \cite{zhang2021beyond}, MIRNet \cite{zamir2020learning}, EnGAN \cite{jiang2021enlightengan}, DeepUPE \cite{wang2019underexposed}, DeepLPF \cite{moran2020deeplpf}, UEGAN \cite{ni2020towards}, SGM \cite{yang2021sparse}, MIRNet-v2 \cite{zamir2022learning}, and SNR-Aware \cite{xu2022snr}. Note that we load the pre-trained parameters of EnGAN and retrain other methods on the same datasets.

\vspace{0.1cm}
\noindent\textbf{Quantitative comparison.}
In this work, PSNR, SSIM \cite{wang2004image}, the $L_{2}-distance$ in CIELAB color space ($\bigtriangleup E _{ab} $), LPIPS \cite{zhang2018unreasonable}, and Color-Sensitive Error (CSE) \cite{zhao2021unsupervised} are employed as the evaluation matrices. Note that LPIPS measures the distance of high-level features between two images, and CSE measures the color difference between two images. We employ the ratio of our result as the unit of CSE following recent works \cite{zhang2022deep,zhao2022fcl}. In general, the lower LPIPS, lower CSE, lower $\bigtriangleup E _{ab} $, higher SSIM, and higher PSNR values represent two images with more relative low-/high- level features. Table \ref{tab:experiments} shows the quantitative comparisons of LOL-real and FiveK. The proposed method achieves the best results on all metrics benefiting from the decoupled strategy. It demonstrates that enhanced images of our method are closer to ground-truth no matter in structure, color, or high-level features. Besides, we also present the comparison of model size. As shown in Table \ref{tab:experiments}, the size of the proposed method is comparable to other methods.

\vspace{0.1cm}
\noindent\textbf{Qualitative comparison.}
We present the visual results of two datasets in Fig. \ref{fig:5k_lol} for comparing the proposed method with some baselines good at PSNR. It can be seen that although existing methods reach decent lightness, they are hard to recover chrominance well (e.g., the color of the sky in Fig. \ref{fig:5k_lol}). While the proposed method can obtain satisfying results. Besides, the proposed method achieves customized enhancement as shown in Fig. \ref{fig:diver_style} and Fig. \ref{fig:colorful}. The users can enhance low-light images as per their preferences.

\subsection{Comparison with Two-Step ``Brighten-and-Colorize'' Methods}
We also compare BCNet with other two-step ``brighten-and-colorize'' methods, which are composed of a brightener (brightening network) and a colorizer (colorization network). The brightener is an LLIE method, whose input and output are single-channel lightness. The colorizer is a user-guide image colorization method, which takes the low-light image as color hints and predicts chrominance based on the output of brightener and color hints. The implementation details of the two-step ``brighten-and-colorize'' can refer to Fig. \ref{fig:b-and-c}. For brighteners, we choose SNR-Aware \cite{xu2022snr} and MIRNet-v2\cite{zamir2022learning}. For colorizers, we choose an exemplar-based Yin \emph{et al.} \cite{yin2021yes} method and a stroke-based Zhang \emph{et al.} \cite{zhang2017real} method. Then, we pair them up to get four methods. Note that the brighteners need to be retrained on our datasets since the mismatched input/output channels of networks. The stroke-based colorizer is easy to be retrained as well. While for the exemplar-based colorizer, it is hard to be retrained in traditional LLIE datasets, since it needs to calculate similarities between the input image and the exemplar image, and query color from the database when lacking matched colors in the exemplar image. Therefore, we opt to solely retrain the stroke-based colorizer,

\noindent while utilizing the pre-trained exemplar-based colorizer.

\begin{figure}
  \centering
  \captionsetup[subfloat]{labelformat=empty}

  \subfloat{
    %\hspace{0.2in}
    \includegraphics[width= 0.9\columnwidth]{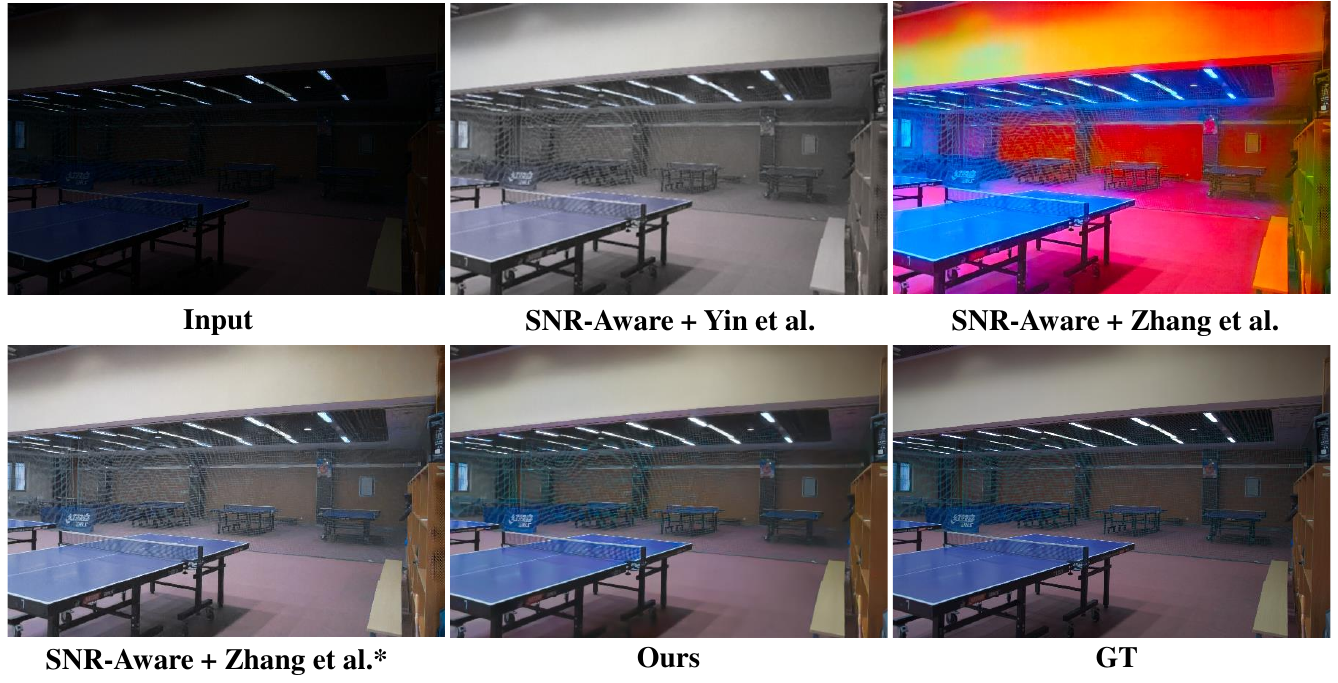}
  }
\vspace{-0.2cm}
  \caption{The qualitative comparison with ``brighten-and-colorize'' methods. Note that two brighteners have similar visual results, we only present the results of SNR-Aware \cite{xu2022snr}. }
    \label{fig:comparison_brighten_and_colorize}
    \vspace{-0.5cm}
    
\end{figure}

The quantitative comparison is presented in Table \ref{tab:brighten-and-colorize-methods}. We adopt PSNR, SSIM, and CSE as the assessment metrics, and it can be seen that the proposed method obtains the best performance. Fig. \ref{fig:comparison_brighten_and_colorize} shows the qualitative results. For the exemplar-based colorizer, which aims to transfer color from the exemplar image to the input lightness, the generated color is unsaturated due to the unsaturated exemplar image (i.e., input low-light image). On the other hand, color hints in stroke-based colorization are just provided as approximate guidance. Although the input color is unsaturated, the generated color of the stroke-based colorizer is bright. After retraining, the generated color can be more realistic. For example, the input color of the Ping-Pong table in Fig. \ref{fig:comparison_brighten_and_colorize} is dull blue, therefore, the result of exemplar-based colorizer is still dull blue, while stroke-based colorizer can generate bright blue and retrained stroke-based colorizer can generate more realistic bright blue. However, the result of the proposed method is more accurate and saturated.
\vspace{-0.3cm}
\subsection{Analysis}

% lol2 in brighten-and-colorize
\begin{table}[]
\caption{Quantitative comparison with two-step ``brighten-and-colorize'' methods on LOL-real \cite{yang2021sparse} dataset.
The best results are boldfaced and the second-best ones are
underlined. Note that ``*'' represents this method is retrained.}
 \vspace{-0.25cm}
    \label{tab:brighten-and-colorize-methods}
    \resizebox{0.9\columnwidth}{!}{
\renewcommand{\arraystretch}{1} % 可以调节, 1.2指高度是默认的1.2倍
    \centering

    \begin{tabular}{c c c c c }
    % \Xhline{1-1}
     \toprule
      Brightener& Colorizer & PSNR $\uparrow$    &   SSIM $\uparrow$ &
CSE(ratio) $\downarrow$ \\
     % \cmidrule(r){1-1} \cmidrule(r){2-2} \cmidrule(r){3-5}
     \midrule
     \multirow{3}{*}{SNR-Aware \cite{xu2022snr}} & Zhang \emph{et al.}* \cite{zhang2017real} &20.90&\underline{0.8304}&1.34 \\
      & Zhang \emph{et al.} \cite{zhang2017real}&11.81&0.5411&70.44\\
     & Yin \emph{et al.} \cite{yin2021yes}&19.69&0.7881&7.40\\
      \midrule
      % \cmidrule(r){1-2}
     \multirow{3}{*}{MIRNet-v2 \cite{zamir2022learning}} & Zhang \emph{et al.}* \cite{zhang2017real} &\underline{21.91}&0.8242&\underline{1.09} \\
      & Zhang \emph{et al.} \cite{zhang2017real}&12.01&0.5313&77.60\\
     & Yin \emph{et al.} \cite{yin2021yes}&20.44&0.7810&7.30\\
     
     \midrule
     \multicolumn{2}{c}{Ours} & \textbf{23.27} & \textbf{0.8637} & \textbf{1.00} \\
     \bottomrule
    \end{tabular}
    }
\vspace{-0.7cm}
\end{table}

\noindent\textbf{Ablation study.} We conduct the ablation study on five different settings to demonstrate the effectiveness of the proposed designs: (1) ``W/o Decoupling'' adopts an encoder-decoder architecture and predicts enhanced results in RGB color space; (2) ``W/o Sharing'' adopts two-step ``brighten-and-colorize'', where brightener and colorizer are based on the proposed designs; (3) ``W/o LAM'' removes the lightness adjustment module; (4) ``W/o CEM'' removes the color embedding module and adopts concatenation; (5) ``W/o $\mathcal{L}_{q}$'' removes color classification loss.

\begin{figure}
  \centering
  \captionsetup[subfloat]{labelformat=empty}
   \subfloat{
    \includegraphics[width=0.9\columnwidth]{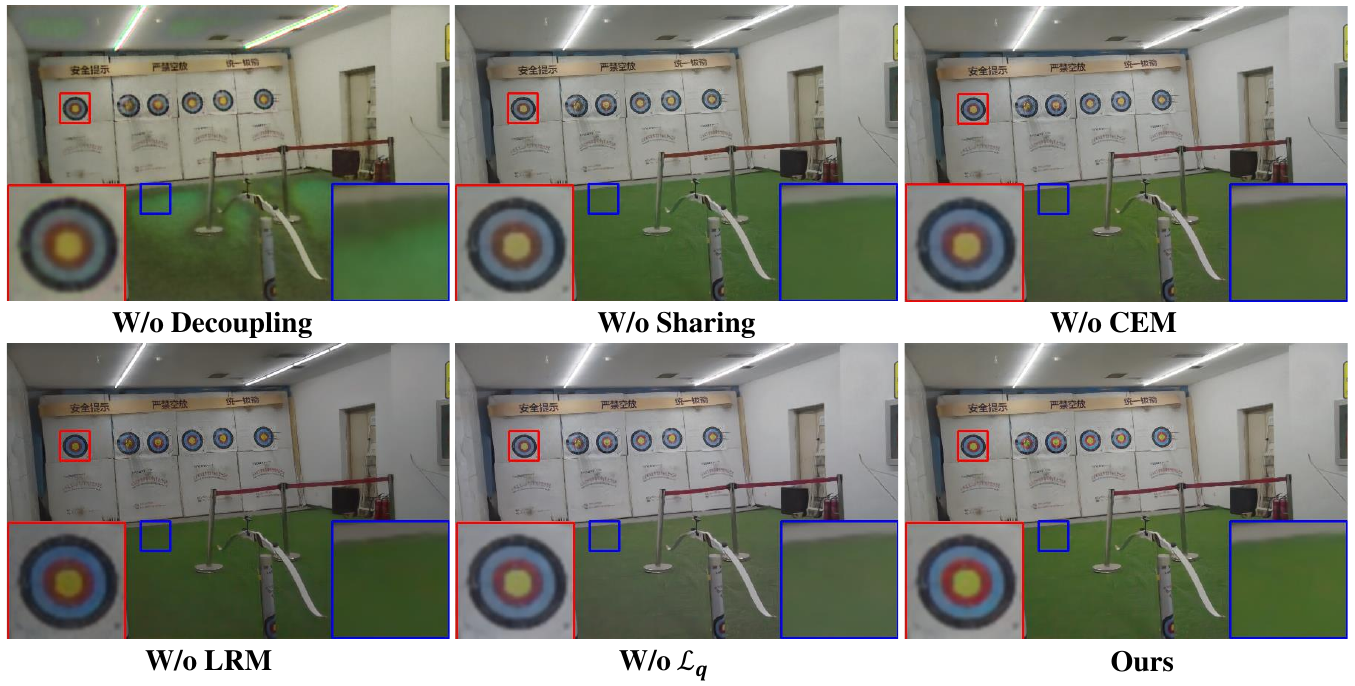}
  }
 \vspace{-0.3cm}
  \caption{The qualitative comparison of ablation studies.}
    \label{fig:ablation}
    \vspace{-0.5cm}
\end{figure}

We report the ablation study results in Table \ref{tab:ablation}. We can see that our full setting yields the best PSNR, SSIM, and second-best CSE. ``W/o Decoupling'' verifies the effectiveness of this decoupled enhancement mode. ``W/o $\mathcal{L}_{q}$'', ``W/o Sharing'', ``W/o LAM'', and ``W/o CEM'' verify the effectiveness of the proposed corresponding designs. Fig. \ref{fig:ablation} presents the visual results of ablation studies. ``W/o $\mathcal{L}_{q}$'', ``W/o Sharing'', and ``W/o CEM'' hardly perform well in chrominance. ``W/o LAM'' may lead to worse performance in lightness. ``W/o Decoupling'' suffers from both above problems. Our result reaches the best visual quality. It is worth mentioning that although the quantitative results of ``W/o $\mathcal{L}_{q}$'' have only a few reductions, the qualitative result suffers from a little color vanishment (e.g., the color of grasses). It verifies the effectiveness of $\mathcal{L}_{q}$. Besides, we also conduct the ablation study in the other loss functions and adopted color spaces, which can be found in the supplementary materials.
% ablation
\begin{table}[]
\caption{Ablation studies on LOL-real \cite{yang2021sparse} dataset.  
The best results
are boldfaced and the second-best ones are underlined.}
 \vspace{-0.25cm}
    \label{tab:ablation}
    \resizebox{0.9\columnwidth}{!}{
\renewcommand{\arraystretch}{1} % 可以调节, 1.2指高度是默认的1.2倍
    \centering
    \begin{tabular}{c c c c c }
        \hline
    Methods& PSNR $\uparrow$    &   SSIM $\uparrow$ &
CSE(ratio) $\downarrow$& Size (M) \\
     \hline
      W/o Decoupling & 21.28 & 0.8003  & 3.05 & 4.22\\
      W/o Sharing & 23.03 & 0.8591 & 1.12 & 8.44 \\
      W/o LAM & 22.77 &  0.8516   & \textbf{0.91} & 6.84 \\
      W/o CEM & 23.06 & 0.8575 &  1.66 & 6.84 \\
      W/o $\mathcal{L}_{q}$ & \underline{23.21} & \underline{0.8600} & 1.11 & 6.84 \\
     \hline
     \multicolumn{1}{c}{Ours} & \textbf{23.27} & \textbf{0.8637}  & \underline{1.00} & 6.84 \\
     \hline
    \end{tabular}}
 \vspace{-0.7cm}
\end{table}

\begin{figure}
  \centering
  \captionsetup[subfloat]{labelformat=empty}

\subfloat{
    %\hspace{0.2in}
    \includegraphics[width=0.9\columnwidth]{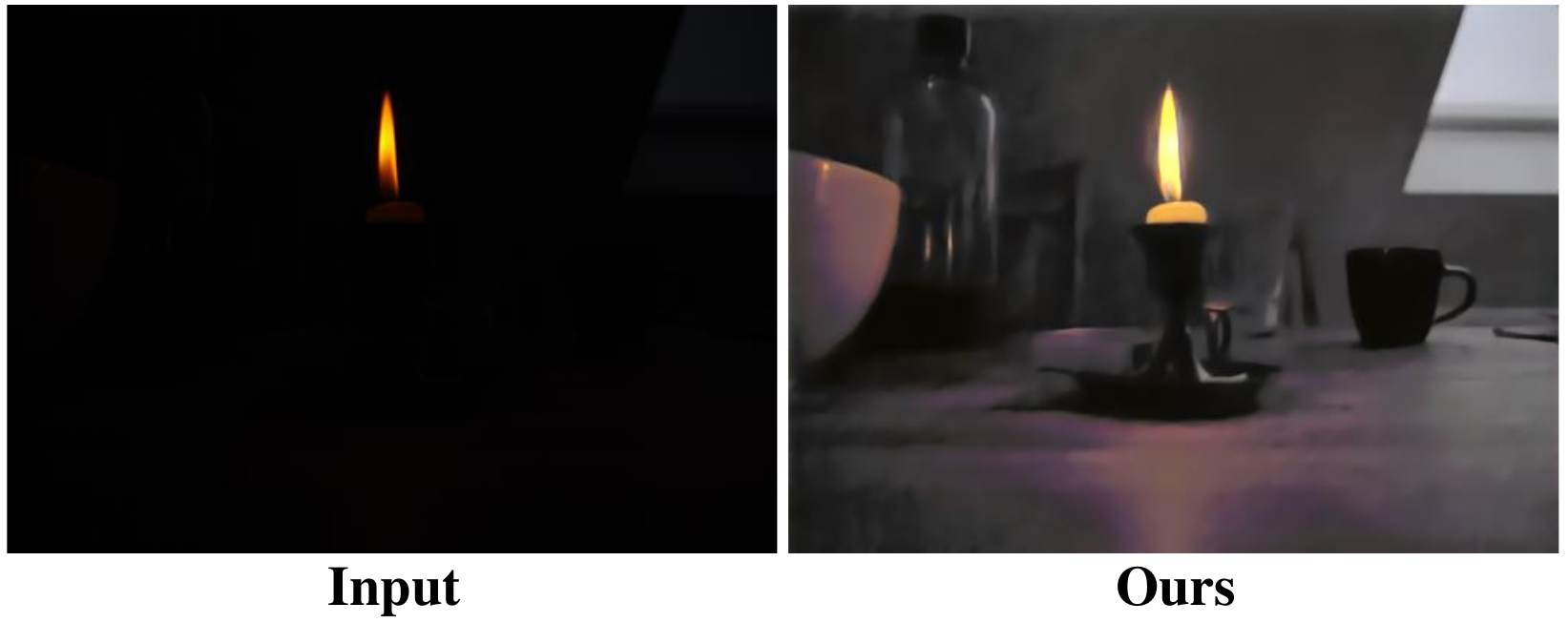}
  }
   \vspace{-0.2cm}
  \caption{Failure case. When the input image is extremely dark, we hardly predict proper chrominance, since the input image hardly provides enough chrominance information.}
    \label{fig:limitation}
    \vspace{-0.3cm}
\end{figure}

\vspace{0.1cm}
\noindent\textbf{Limitation.}  The precondition of the colorization sub-task is that the chrominance of the input image contains a little color tone and shapes. However, when the input image is extremely dark, it hardly provides enough information for colorization. As a result, the chrominance of the enhanced image is very dull as shown in Fig. \ref{fig:limitation}. Actually, it may be relieved by the automatic colorization methods, which is also one of our future works.

\section{Conclusion}
In this work, we novelly introduce image colorization to LLIE and propose a ``brighten-and-colorize'' enhancement network BCNet for low-light images. Based on the relation of image colorization and LLIE, we treat this enhancement as a multi-task learning problem. A low-light image is decomposed to lightness and chrominance and fulfills the decoupled enhancement by the proposed BCNet. BCNet contains a multi-task encoder and two task-specific decoders, which are designed elaborately according to different characteristics of lightness and chrominance. Based on the decoupled design, BCNet further achieves lightness-invariant customization with diverse saturations and color styles by manipulating the color guidance of the colorization sub-task. Extensive experiments demonstrate that BCNet sets the current SOTA results in LLIE and flexible customization. In the future, we will explore more effective ways to introduce image colorization to other image restoration tasks.

\begin{acks}
This work was supported by the National Natural Science Foundation of China under Grant No. 62071500. Supported by Sino-Germen Mobility Programme M-0421. 
\end{acks}

\balance
\bibliographystyle{ACM-Reference-Format}
\bibliography{sample-base}


\begin{thebibliography}{57}

%%% the Web address.

\ifx \showCODEN    \undefined \def \showCODEN     #1{\unskip}     \fi
\ifx \showDOI      \undefined \def \showDOI       #1{#1}\fi
\ifx \showISBNx    \undefined \def \showISBNx     #1{\unskip}     \fi
\ifx \showISBNxiii \undefined \def \showISBNxiii  #1{\unskip}     \fi
\ifx \showISSN     \undefined \def \showISSN      #1{\unskip}     \fi
\ifx \showLCCN     \undefined \def \showLCCN      #1{\unskip}     \fi
\ifx \shownote     \undefined \def \shownote      #1{#1}          \fi
\ifx \showarticletitle \undefined \def \showarticletitle #1{#1}   \fi
\ifx \showURL      \undefined \def \showURL       {\relax}        \fi
% invisible to TeX
\providecommand\bibfield[2]{#2}
\providecommand\bibinfo[2]{#2}
\providecommand\natexlab[1]{#1}
\providecommand\showeprint[2][]{arXiv:#2}

\bibitem[Bychkovsky et~al\mbox{.}(2011)]%
        {bychkovsky2011learning}
\bibfield{author}{\bibinfo{person}{Vladimir Bychkovsky},
  \bibinfo{person}{Sylvain Paris}, \bibinfo{person}{Eric Chan}, {and}
  \bibinfo{person}{Fr{\'e}do Durand}.} \bibinfo{year}{2011}\natexlab{}.
\newblock \showarticletitle{Learning photographic global tonal adjustment with
  a database of input/output image pairs}. In \bibinfo{booktitle}{\emph{IEEE
  Conf. Comput. Vis. Pattern Recog.}} IEEE, \bibinfo{pages}{97--104}.
\newblock


\bibitem[Chambolle(2004)]%
        {chambolle2004algorithm}
\bibfield{author}{\bibinfo{person}{Antonin Chambolle}.}
  \bibinfo{year}{2004}\natexlab{}.
\newblock \showarticletitle{An algorithm for total variation minimization and
  applications}.
\newblock \bibinfo{journal}{\emph{Journal of Mathematical Imaging and Vision}}
  \bibinfo{volume}{20} (\bibinfo{year}{2004}), \bibinfo{pages}{89--97}.
\newblock


\bibitem[Gharbi et~al\mbox{.}(2017)]%
        {gharbi2017deep}
\bibfield{author}{\bibinfo{person}{Micha{\"e}l Gharbi}, \bibinfo{person}{Jiawen
  Chen}, \bibinfo{person}{Jonathan~T Barron}, \bibinfo{person}{Samuel~W
  Hasinoff}, {and} \bibinfo{person}{Fr{\'e}do Durand}.}
  \bibinfo{year}{2017}\natexlab{}.
\newblock \showarticletitle{Deep bilateral learning for real-time image
  enhancement}.
\newblock \bibinfo{journal}{\emph{ACM Trans. Graph.}} \bibinfo{volume}{36},
  \bibinfo{number}{4} (\bibinfo{year}{2017}), \bibinfo{pages}{1--12}.
\newblock


\bibitem[Guadarrama et~al\mbox{.}(2017)]%
        {guadarrama2017pixcolor}
\bibfield{author}{\bibinfo{person}{Sergio Guadarrama}, \bibinfo{person}{Ryan
  Dahl}, \bibinfo{person}{David Bieber}, \bibinfo{person}{Mohammad Norouzi},
  \bibinfo{person}{Jonathon Shlens}, {and} \bibinfo{person}{Kevin Murphy}.}
  \bibinfo{year}{2017}\natexlab{}.
\newblock \showarticletitle{Pixcolor: Pixel recursive colorization}.
\newblock \bibinfo{journal}{\emph{arXiv preprint arXiv:1705.07208}}
  (\bibinfo{year}{2017}).
\newblock


\bibitem[Guo et~al\mbox{.}(2020)]%
        {guo2020zero}
\bibfield{author}{\bibinfo{person}{Chunle Guo}, \bibinfo{person}{Chongyi Li},
  \bibinfo{person}{Jichang Guo}, \bibinfo{person}{Chen~Change Loy},
  \bibinfo{person}{Junhui Hou}, \bibinfo{person}{Sam Kwong}, {and}
  \bibinfo{person}{Runmin Cong}.} \bibinfo{year}{2020}\natexlab{}.
\newblock \showarticletitle{Zero-reference deep curve estimation for low-light
  image enhancement}. In \bibinfo{booktitle}{\emph{IEEE Conf. Comput. Vis.
  Pattern Recog.}} \bibinfo{pages}{1780--1789}.
\newblock


\bibitem[He et~al\mbox{.}(2018)]%
        {he2018deep}
\bibfield{author}{\bibinfo{person}{Mingming He}, \bibinfo{person}{Dongdong
  Chen}, \bibinfo{person}{Jing Liao}, \bibinfo{person}{Pedro~V Sander}, {and}
  \bibinfo{person}{Lu Yuan}.} \bibinfo{year}{2018}\natexlab{}.
\newblock \showarticletitle{Deep exemplar-based colorization}.
\newblock \bibinfo{journal}{\emph{ACM Trans. Graph.}} \bibinfo{volume}{37},
  \bibinfo{number}{4} (\bibinfo{year}{2018}), \bibinfo{pages}{1--16}.
\newblock


\bibitem[Huang et~al\mbox{.}(2022)]%
        {huang2022deep}
\bibfield{author}{\bibinfo{person}{Jie Huang}, \bibinfo{person}{Yajing Liu},
  \bibinfo{person}{Feng Zhao}, \bibinfo{person}{Keyu Yan},
  \bibinfo{person}{Jinghao Zhang}, \bibinfo{person}{Yukun Huang},
  \bibinfo{person}{Man Zhou}, {and} \bibinfo{person}{Zhiwei Xiong}.}
  \bibinfo{year}{2022}\natexlab{}.
\newblock \showarticletitle{Deep Fourier-Based Exposure Correction Network with
  Spatial-Frequency Interaction}. In \bibinfo{booktitle}{\emph{Eur. Conf.
  Comput. Vis.}} Springer, \bibinfo{pages}{163--180}.
\newblock


\bibitem[Jiang et~al\mbox{.}(2021)]%
        {jiang2021enlightengan}
\bibfield{author}{\bibinfo{person}{Yifan Jiang}, \bibinfo{person}{Xinyu Gong},
  \bibinfo{person}{Ding Liu}, \bibinfo{person}{Yu Cheng}, \bibinfo{person}{Chen
  Fang}, \bibinfo{person}{Xiaohui Shen}, \bibinfo{person}{Jianchao Yang},
  \bibinfo{person}{Pan Zhou}, {and} \bibinfo{person}{Zhangyang Wang}.}
  \bibinfo{year}{2021}\natexlab{}.
\newblock \showarticletitle{Enlightengan: Deep light enhancement without paired
  supervision}.
\newblock \bibinfo{journal}{\emph{IEEE Trans. Image Process.}}
  \bibinfo{volume}{30} (\bibinfo{year}{2021}), \bibinfo{pages}{2340--2349}.
\newblock


\bibitem[Jin et~al\mbox{.}(2019)]%
        {jin2019flexible}
\bibfield{author}{\bibinfo{person}{Zhi Jin}, \bibinfo{person}{Muhammad~Zafar
  Iqbal}, \bibinfo{person}{Dmytro Bobkov}, \bibinfo{person}{Wenbin Zou},
  \bibinfo{person}{Xia Li}, {and} \bibinfo{person}{Eckehard Steinbach}.}
  \bibinfo{year}{2019}\natexlab{}.
\newblock \showarticletitle{A flexible deep CNN framework for image
  restoration}.
\newblock \bibinfo{journal}{\emph{IEEE Trans. Multimedia}}
  \bibinfo{volume}{22}, \bibinfo{number}{4} (\bibinfo{year}{2019}),
  \bibinfo{pages}{1055--1068}.
\newblock


\bibitem[Jin et~al\mbox{.}(2020)]%
        {jin2020dual}
\bibfield{author}{\bibinfo{person}{Zhi Jin}, \bibinfo{person}{Muhammad~Zafar
  Iqbal}, \bibinfo{person}{Wenbin Zou}, \bibinfo{person}{Xia Li}, {and}
  \bibinfo{person}{Eckehard Steinbach}.} \bibinfo{year}{2020}\natexlab{}.
\newblock \showarticletitle{Dual-stream multi-path recursive residual network
  for JPEG image compression artifacts reduction}.
\newblock \bibinfo{journal}{\emph{IEEE Trans. Circuit Syst. Video Technol.}}
  \bibinfo{volume}{31}, \bibinfo{number}{2} (\bibinfo{year}{2020}),
  \bibinfo{pages}{467--479}.
\newblock


\bibitem[Jobson et~al\mbox{.}(1997)]%
        {jobson1997multiscale}
\bibfield{author}{\bibinfo{person}{Daniel~J Jobson}, \bibinfo{person}{Zia-ur
  Rahman}, {and} \bibinfo{person}{Glenn~A Woodell}.}
  \bibinfo{year}{1997}\natexlab{}.
\newblock \showarticletitle{A multiscale retinex for bridging the gap between
  color images and the human observation of scenes}.
\newblock \bibinfo{journal}{\emph{IEEE Trans. Image Process.}}
  \bibinfo{volume}{6}, \bibinfo{number}{7} (\bibinfo{year}{1997}),
  \bibinfo{pages}{965--976}.
\newblock


\bibitem[Kim et~al\mbox{.}(2021)]%
        {kim2021representative}
\bibfield{author}{\bibinfo{person}{Hanul Kim}, \bibinfo{person}{Su-Min Choi},
  \bibinfo{person}{Chang-Su Kim}, {and} \bibinfo{person}{Yeong~Jun Koh}.}
  \bibinfo{year}{2021}\natexlab{}.
\newblock \showarticletitle{Representative color transform for image
  enhancement}. In \bibinfo{booktitle}{\emph{Int. Conf. Comput. Vis.}}
  \bibinfo{pages}{4459--4468}.
\newblock


\bibitem[Kim et~al\mbox{.}(2020a)]%
        {kim2020global}
\bibfield{author}{\bibinfo{person}{Han-Ul Kim}, \bibinfo{person}{Young~Jun
  Koh}, {and} \bibinfo{person}{Chang-Su Kim}.}
  \bibinfo{year}{2020}\natexlab{a}.
\newblock \showarticletitle{Global and local enhancement networks for paired
  and unpaired image enhancement}. In \bibinfo{booktitle}{\emph{Eur. Conf.
  Comput. Vis.}} Springer, \bibinfo{pages}{339--354}.
\newblock


\bibitem[Kim et~al\mbox{.}(2020b)]%
        {kim2020pienet}
\bibfield{author}{\bibinfo{person}{Han-Ul Kim}, \bibinfo{person}{Young~Jun
  Koh}, {and} \bibinfo{person}{Chang-Su Kim}.}
  \bibinfo{year}{2020}\natexlab{b}.
\newblock \showarticletitle{PieNet: Personalized image enhancement network}. In
  \bibinfo{booktitle}{\emph{Eur. Conf. Comput. Vis.}} Springer,
  \bibinfo{pages}{374--390}.
\newblock


\bibitem[Kingma and Ba(2015)]%
        {kingma2014adam}
\bibfield{author}{\bibinfo{person}{Diederik~P Kingma} {and}
  \bibinfo{person}{Jimmy Ba}.} \bibinfo{year}{2015}\natexlab{}.
\newblock \showarticletitle{Adam: A Method for Stochastic Optimization}. In
  \bibinfo{booktitle}{\emph{Int. Conf. Learn. Represent.}}
\newblock


\bibitem[Kingma and Dhariwal(2018)]%
        {kingma2018glow}
\bibfield{author}{\bibinfo{person}{Durk~P Kingma} {and}
  \bibinfo{person}{Prafulla Dhariwal}.} \bibinfo{year}{2018}\natexlab{}.
\newblock \showarticletitle{Glow: Generative flow with invertible 1x1
  convolutions}.
\newblock \bibinfo{journal}{\emph{Adv. Neural Inform. Process. Syst.}}
  \bibinfo{volume}{31} (\bibinfo{year}{2018}).
\newblock


\bibitem[Kumar et~al\mbox{.}(2020)]%
        {kumar2021colorization}
\bibfield{author}{\bibinfo{person}{Manoj Kumar}, \bibinfo{person}{Dirk
  Weissenborn}, {and} \bibinfo{person}{Nal Kalchbrenner}.}
  \bibinfo{year}{2020}\natexlab{}.
\newblock \showarticletitle{Colorization Transformer}. In
  \bibinfo{booktitle}{\emph{Int. Conf. Learn. Represent.}}
\newblock


\bibitem[Lai et~al\mbox{.}(2018)]%
        {lai2018fast}
\bibfield{author}{\bibinfo{person}{Wei-Sheng Lai}, \bibinfo{person}{Jia-Bin
  Huang}, \bibinfo{person}{Narendra Ahuja}, {and} \bibinfo{person}{Ming-Hsuan
  Yang}.} \bibinfo{year}{2018}\natexlab{}.
\newblock \showarticletitle{Fast and accurate image super-resolution with deep
  laplacian pyramid networks}.
\newblock \bibinfo{journal}{\emph{IEEE Trans. Pattern Anal. Mach. Intell.}}
  \bibinfo{volume}{41}, \bibinfo{number}{11} (\bibinfo{year}{2018}),
  \bibinfo{pages}{2599--2613}.
\newblock


\bibitem[Li et~al\mbox{.}(2023)]%
        {li2023embedding}
\bibfield{author}{\bibinfo{person}{Chongyi Li}, \bibinfo{person}{Chun-Le Guo},
  \bibinfo{person}{Man Zhou}, \bibinfo{person}{Zhexin Liang},
  \bibinfo{person}{Shangchen Zhou}, \bibinfo{person}{Ruicheng Feng}, {and}
  \bibinfo{person}{Chen~Change Loy}.} \bibinfo{year}{2023}\natexlab{}.
\newblock \showarticletitle{Embedding Fourier for Ultra-High-Definition
  Low-Light Image Enhancement}.
\newblock \bibinfo{journal}{\emph{arXiv preprint arXiv:2302.11831}}
  (\bibinfo{year}{2023}).
\newblock


\bibitem[Li et~al\mbox{.}(2019)]%
        {li2019selective}
\bibfield{author}{\bibinfo{person}{Xiang Li}, \bibinfo{person}{Wenhai Wang},
  \bibinfo{person}{Xiaolin Hu}, {and} \bibinfo{person}{Jian Yang}.}
  \bibinfo{year}{2019}\natexlab{}.
\newblock \showarticletitle{Selective kernel networks}. In
  \bibinfo{booktitle}{\emph{IEEE Conf. Comput. Vis. Pattern Recog.}}
  \bibinfo{pages}{510--519}.
\newblock


\bibitem[Liu et~al\mbox{.}(2021)]%
        {liu2021retinex}
\bibfield{author}{\bibinfo{person}{Risheng Liu}, \bibinfo{person}{Long Ma},
  \bibinfo{person}{Jiaao Zhang}, \bibinfo{person}{Xin Fan}, {and}
  \bibinfo{person}{Zhongxuan Luo}.} \bibinfo{year}{2021}\natexlab{}.
\newblock \showarticletitle{Retinex-inspired unrolling with cooperative prior
  architecture search for low-light image enhancement}. In
  \bibinfo{booktitle}{\emph{IEEE Conf. Comput. Vis. Pattern Recog.}}
  \bibinfo{pages}{10561--10570}.
\newblock


\bibitem[Lore et~al\mbox{.}(2017)]%
        {lore2017llnet}
\bibfield{author}{\bibinfo{person}{Kin~Gwn Lore}, \bibinfo{person}{Adedotun
  Akintayo}, {and} \bibinfo{person}{Soumik Sarkar}.}
  \bibinfo{year}{2017}\natexlab{}.
\newblock \showarticletitle{LLNet: A deep autoencoder approach to natural
  low-light image enhancement}.
\newblock \bibinfo{journal}{\emph{Pattern Recognition}}  \bibinfo{volume}{61}
  (\bibinfo{year}{2017}), \bibinfo{pages}{650--662}.
\newblock


\bibitem[Lu et~al\mbox{.}(2020)]%
        {lu2020gray2colornet}
\bibfield{author}{\bibinfo{person}{Peng Lu}, \bibinfo{person}{Jinbei Yu},
  \bibinfo{person}{Xujun Peng}, \bibinfo{person}{Zhaoran Zhao}, {and}
  \bibinfo{person}{Xiaojie Wang}.} \bibinfo{year}{2020}\natexlab{}.
\newblock \showarticletitle{Gray2colornet: Transfer more colors from reference
  image}. In \bibinfo{booktitle}{\emph{ACM Int. Conf. Multimedia}}.
  \bibinfo{pages}{3210--3218}.
\newblock


\bibitem[Ma et~al\mbox{.}(2022)]%
        {ma2022toward}
\bibfield{author}{\bibinfo{person}{Long Ma}, \bibinfo{person}{Tengyu Ma},
  \bibinfo{person}{Risheng Liu}, \bibinfo{person}{Xin Fan}, {and}
  \bibinfo{person}{Zhongxuan Luo}.} \bibinfo{year}{2022}\natexlab{}.
\newblock \showarticletitle{Toward fast, flexible, and robust low-light image
  enhancement}. In \bibinfo{booktitle}{\emph{IEEE Conf. Comput. Vis. Pattern
  Recog.}} \bibinfo{pages}{5637--5646}.
\newblock


\bibitem[Moran et~al\mbox{.}(2020)]%
        {moran2020deeplpf}
\bibfield{author}{\bibinfo{person}{Sean Moran}, \bibinfo{person}{Pierre Marza},
  \bibinfo{person}{Steven McDonagh}, \bibinfo{person}{Sarah Parisot}, {and}
  \bibinfo{person}{Gregory Slabaugh}.} \bibinfo{year}{2020}\natexlab{}.
\newblock \showarticletitle{Deeplpf: Deep local parametric filters for image
  enhancement}. In \bibinfo{booktitle}{\emph{IEEE Conf. Comput. Vis. Pattern
  Recog.}} \bibinfo{pages}{12826--12835}.
\newblock


\bibitem[Ni et~al\mbox{.}(2020)]%
        {ni2020towards}
\bibfield{author}{\bibinfo{person}{Zhangkai Ni}, \bibinfo{person}{Wenhan Yang},
  \bibinfo{person}{Shiqi Wang}, \bibinfo{person}{Lin Ma}, {and}
  \bibinfo{person}{Sam Kwong}.} \bibinfo{year}{2020}\natexlab{}.
\newblock \showarticletitle{Towards unsupervised deep image enhancement with
  generative adversarial network}.
\newblock \bibinfo{journal}{\emph{IEEE Trans. Image Process.}}
  \bibinfo{volume}{29} (\bibinfo{year}{2020}), \bibinfo{pages}{9140--9151}.
\newblock


\bibitem[Pan et~al\mbox{.}(2021)]%
        {pan2021view}
\bibfield{author}{\bibinfo{person}{Qingzhe Pan}, \bibinfo{person}{Zhifu Zhao},
  \bibinfo{person}{Xuemei Xie}, \bibinfo{person}{Jianan Li},
  \bibinfo{person}{Yuhan Cao}, {and} \bibinfo{person}{Guangming Shi}.}
  \bibinfo{year}{2021}\natexlab{}.
\newblock \showarticletitle{View-normalized skeleton generation for action
  recognition}. In \bibinfo{booktitle}{\emph{ACM Int. Conf. Multimedia}}.
  \bibinfo{pages}{1875--1883}.
\newblock


\bibitem[Pizer et~al\mbox{.}(1987)]%
        {pizer1987adaptive}
\bibfield{author}{\bibinfo{person}{Stephen~M Pizer}, \bibinfo{person}{E~Philip
  Amburn}, \bibinfo{person}{John~D Austin}, \bibinfo{person}{Robert Cromartie},
  \bibinfo{person}{Ari Geselowitz}, \bibinfo{person}{Trey Greer},
  \bibinfo{person}{Bart ter Haar~Romeny}, \bibinfo{person}{John~B Zimmerman},
  {and} \bibinfo{person}{Karel Zuiderveld}.} \bibinfo{year}{1987}\natexlab{}.
\newblock \showarticletitle{Adaptive histogram equalization and its
  variations}.
\newblock \bibinfo{journal}{\emph{Computer vision, graphics, and image
  processing}} \bibinfo{volume}{39}, \bibinfo{number}{3}
  (\bibinfo{year}{1987}), \bibinfo{pages}{355--368}.
\newblock


\bibitem[Rahman et~al\mbox{.}(2004)]%
        {rahman2004retinex}
\bibfield{author}{\bibinfo{person}{Zia-ur Rahman}, \bibinfo{person}{Daniel~J
  Jobson}, {and} \bibinfo{person}{Glenn~A Woodell}.}
  \bibinfo{year}{2004}\natexlab{}.
\newblock \showarticletitle{Retinex processing for automatic image
  enhancement}.
\newblock \bibinfo{journal}{\emph{Journal of Electronic imaging}}
  \bibinfo{volume}{13}, \bibinfo{number}{1} (\bibinfo{year}{2004}),
  \bibinfo{pages}{100--110}.
\newblock


\bibitem[Reinhard et~al\mbox{.}(2001)]%
        {reinhard2001color}
\bibfield{author}{\bibinfo{person}{Erik Reinhard}, \bibinfo{person}{Michael
  Adhikhmin}, \bibinfo{person}{Bruce Gooch}, {and} \bibinfo{person}{Peter
  Shirley}.} \bibinfo{year}{2001}\natexlab{}.
\newblock \showarticletitle{Color transfer between images}.
\newblock \bibinfo{journal}{\emph{IEEE Computer Graphics and Applications}}
  \bibinfo{volume}{21}, \bibinfo{number}{5} (\bibinfo{year}{2001}),
  \bibinfo{pages}{34--41}.
\newblock


\bibitem[Ren et~al\mbox{.}(2015)]%
        {ren2015faster}
\bibfield{author}{\bibinfo{person}{Shaoqing Ren}, \bibinfo{person}{Kaiming He},
  \bibinfo{person}{Ross Girshick}, {and} \bibinfo{person}{Jian Sun}.}
  \bibinfo{year}{2015}\natexlab{}.
\newblock \showarticletitle{Faster r-cnn: Towards real-time object detection
  with region proposal networks}.
\newblock \bibinfo{journal}{\emph{Adv. Neural Inform. Process. Syst.}}
  \bibinfo{volume}{28} (\bibinfo{year}{2015}).
\newblock


\bibitem[Simonyan and Zisserman(2014)]%
        {simonyan2014very}
\bibfield{author}{\bibinfo{person}{Karen Simonyan} {and}
  \bibinfo{person}{Andrew Zisserman}.} \bibinfo{year}{2014}\natexlab{}.
\newblock \showarticletitle{Very deep convolutional networks for large-scale
  image recognition}.
\newblock \bibinfo{journal}{\emph{arXiv preprint arXiv:1409.1556}}
  (\bibinfo{year}{2014}).
\newblock


\bibitem[Su et~al\mbox{.}(2020)]%
        {su2020instance}
\bibfield{author}{\bibinfo{person}{Jheng-Wei Su}, \bibinfo{person}{Hung-Kuo
  Chu}, {and} \bibinfo{person}{Jia-Bin Huang}.}
  \bibinfo{year}{2020}\natexlab{}.
\newblock \showarticletitle{Instance-aware image colorization}. In
  \bibinfo{booktitle}{\emph{IEEE Conf. Comput. Vis. Pattern Recog.}}
  \bibinfo{pages}{7968--7977}.
\newblock


\bibitem[Sun et~al\mbox{.}(2021)]%
        {sun2021enhance}
\bibfield{author}{\bibinfo{person}{Xiaopeng Sun}, \bibinfo{person}{Muxingzi
  Li}, \bibinfo{person}{Tianyu He}, {and} \bibinfo{person}{Lubin Fan}.}
  \bibinfo{year}{2021}\natexlab{}.
\newblock \showarticletitle{Enhance images as you like with unpaired learning}.
\newblock \bibinfo{journal}{\emph{arXiv preprint arXiv:2110.01161}}
  (\bibinfo{year}{2021}).
\newblock


\bibitem[Vaswani et~al\mbox{.}(2017)]%
        {vaswani2017attention}
\bibfield{author}{\bibinfo{person}{Ashish Vaswani}, \bibinfo{person}{Noam
  Shazeer}, \bibinfo{person}{Niki Parmar}, \bibinfo{person}{Jakob Uszkoreit},
  \bibinfo{person}{Llion Jones}, \bibinfo{person}{Aidan~N Gomez},
  \bibinfo{person}{{\L}ukasz Kaiser}, {and} \bibinfo{person}{Illia
  Polosukhin}.} \bibinfo{year}{2017}\natexlab{}.
\newblock \showarticletitle{Attention is all you need}.
\newblock \bibinfo{journal}{\emph{Adv. Neural Inform. Process. Syst.}}
  \bibinfo{volume}{30} (\bibinfo{year}{2017}).
\newblock


\bibitem[Wang et~al\mbox{.}(2022)]%
        {wang2022learning}
\bibfield{author}{\bibinfo{person}{Haolin Wang}, \bibinfo{person}{Jiawei
  Zhang}, \bibinfo{person}{Ming Liu}, \bibinfo{person}{Xiaohe Wu}, {and}
  \bibinfo{person}{Wangmeng Zuo}.} \bibinfo{year}{2022}\natexlab{}.
\newblock \showarticletitle{Learning Diverse Tone Styles for Image Retouching}.
\newblock \bibinfo{journal}{\emph{arXiv preprint arXiv:2207.05430}}
  (\bibinfo{year}{2022}).
\newblock


\bibitem[Wang et~al\mbox{.}(2019)]%
        {wang2019underexposed}
\bibfield{author}{\bibinfo{person}{Ruixing Wang}, \bibinfo{person}{Qing Zhang},
  \bibinfo{person}{Chi-Wing Fu}, \bibinfo{person}{Xiaoyong Shen},
  \bibinfo{person}{Wei-Shi Zheng}, {and} \bibinfo{person}{Jiaya Jia}.}
  \bibinfo{year}{2019}\natexlab{}.
\newblock \showarticletitle{Underexposed photo enhancement using deep
  illumination estimation}. In \bibinfo{booktitle}{\emph{IEEE Conf. Comput.
  Vis. Pattern Recog.}} \bibinfo{pages}{6849--6857}.
\newblock


\bibitem[Wang et~al\mbox{.}(2004)]%
        {wang2004image}
\bibfield{author}{\bibinfo{person}{Zhou Wang}, \bibinfo{person}{Alan~C Bovik},
  \bibinfo{person}{Hamid~R Sheikh}, {and} \bibinfo{person}{Eero~P Simoncelli}.}
  \bibinfo{year}{2004}\natexlab{}.
\newblock \showarticletitle{Image quality assessment: from error visibility to
  structural similarity}.
\newblock \bibinfo{journal}{\emph{IEEE Trans. Image Process.}}
  \bibinfo{volume}{13}, \bibinfo{number}{4} (\bibinfo{year}{2004}),
  \bibinfo{pages}{600--612}.
\newblock


\bibitem[Wei et~al\mbox{.}(2018)]%
        {wei2018deep}
\bibfield{author}{\bibinfo{person}{Chen Wei}, \bibinfo{person}{Wenjing Wang},
  \bibinfo{person}{Wenhan Yang}, {and} \bibinfo{person}{Jiaying Liu}.}
  \bibinfo{year}{2018}\natexlab{}.
\newblock \showarticletitle{Deep retinex decomposition for low-light
  enhancement}. In \bibinfo{booktitle}{\emph{Brit. Mach. Vis. Conf.}}
\newblock


\bibitem[Wu et~al\mbox{.}(2022)]%
        {wu2022lightweight}
\bibfield{author}{\bibinfo{person}{Hongjun Wu}, \bibinfo{person}{Haoran Qi},
  \bibinfo{person}{Jingzhou Luo}, \bibinfo{person}{Yining Li}, {and}
  \bibinfo{person}{Zhi Jin}.} \bibinfo{year}{2022}\natexlab{}.
\newblock \showarticletitle{A Lightweight Image Entropy-Based
  Divide-and-Conquer Network for Low-Light Image Enhancement}. In
  \bibinfo{booktitle}{\emph{Int. Conf. Multimedia and Expo}}. IEEE,
  \bibinfo{pages}{01--06}.
\newblock


\bibitem[Xu et~al\mbox{.}(2022)]%
        {xu2022snr}
\bibfield{author}{\bibinfo{person}{Xiaogang Xu}, \bibinfo{person}{Ruixing
  Wang}, \bibinfo{person}{Chi-Wing Fu}, {and} \bibinfo{person}{Jiaya Jia}.}
  \bibinfo{year}{2022}\natexlab{}.
\newblock \showarticletitle{SNR-Aware Low-Light Image Enhancement}. In
  \bibinfo{booktitle}{\emph{IEEE Conf. Comput. Vis. Pattern Recog.}}
  \bibinfo{pages}{17714--17724}.
\newblock


\bibitem[Yang et~al\mbox{.}(2019)]%
        {yang2019joint}
\bibfield{author}{\bibinfo{person}{Bowen Yang}, \bibinfo{person}{Chun Yang},
  \bibinfo{person}{Qi Liu}, {and} \bibinfo{person}{Xu-Cheng Yin}.}
  \bibinfo{year}{2019}\natexlab{}.
\newblock \showarticletitle{Joint rotation-invariance face detection and
  alignment with angle-sensitivity cascaded networks}. In
  \bibinfo{booktitle}{\emph{ACM Int. Conf. Multimedia}}.
  \bibinfo{pages}{1473--1480}.
\newblock


\bibitem[Yang et~al\mbox{.}(2021)]%
        {yang2021sparse}
\bibfield{author}{\bibinfo{person}{Wenhan Yang}, \bibinfo{person}{Wenjing
  Wang}, \bibinfo{person}{Haofeng Huang}, \bibinfo{person}{Shiqi Wang}, {and}
  \bibinfo{person}{Jiaying Liu}.} \bibinfo{year}{2021}\natexlab{}.
\newblock \showarticletitle{Sparse gradient regularized deep retinex network
  for robust low-light image enhancement}.
\newblock \bibinfo{journal}{\emph{IEEE Trans. Image Process.}}
  \bibinfo{volume}{30} (\bibinfo{year}{2021}), \bibinfo{pages}{2072--2086}.
\newblock


\bibitem[Yin et~al\mbox{.}(2021)]%
        {yin2021yes}
\bibfield{author}{\bibinfo{person}{Wang Yin}, \bibinfo{person}{Peng Lu},
  \bibinfo{person}{Zhaoran Zhao}, {and} \bibinfo{person}{Xujun Peng}.}
  \bibinfo{year}{2021}\natexlab{}.
\newblock \showarticletitle{Yes,`` Attention Is All You Need", for Exemplar
  based Colorization}. In \bibinfo{booktitle}{\emph{ACM Int. Conf.
  Multimedia}}. \bibinfo{pages}{2243--2251}.
\newblock


\bibitem[Zamir et~al\mbox{.}(2022)]%
        {zamir2022learning}
\bibfield{author}{\bibinfo{person}{SW Zamir}, \bibinfo{person}{A Arora},
  \bibinfo{person}{SH Khan}, \bibinfo{person}{H Munawar}, \bibinfo{person}{FS
  Khan}, \bibinfo{person}{MH Yang}, {and} \bibinfo{person}{L Shao}.}
  \bibinfo{year}{2022}\natexlab{}.
\newblock \showarticletitle{Learning Enriched Features for Fast Image
  Restoration and Enhancement.}
\newblock \bibinfo{journal}{\emph{IEEE Trans. Pattern Anal. Mach. Intell.}}
  (\bibinfo{year}{2022}).
\newblock


\bibitem[Zamir et~al\mbox{.}(2020)]%
        {zamir2020learning}
\bibfield{author}{\bibinfo{person}{Syed~Waqas Zamir}, \bibinfo{person}{Aditya
  Arora}, \bibinfo{person}{Salman Khan}, \bibinfo{person}{Munawar Hayat},
  \bibinfo{person}{Fahad~Shahbaz Khan}, \bibinfo{person}{Ming-Hsuan Yang},
  {and} \bibinfo{person}{Ling Shao}.} \bibinfo{year}{2020}\natexlab{}.
\newblock \showarticletitle{Learning enriched features for real image
  restoration and enhancement}. In \bibinfo{booktitle}{\emph{Eur. Conf. Comput.
  Vis.}} Springer, \bibinfo{pages}{492--511}.
\newblock


\bibitem[Zhang et~al\mbox{.}(2021a)]%
        {zhang2021rellie}
\bibfield{author}{\bibinfo{person}{Rongkai Zhang}, \bibinfo{person}{Lanqing
  Guo}, \bibinfo{person}{Siyu Huang}, {and} \bibinfo{person}{Bihan Wen}.}
  \bibinfo{year}{2021}\natexlab{a}.
\newblock \showarticletitle{ReLLIE: Deep reinforcement learning for customized
  low-light image enhancement}. In \bibinfo{booktitle}{\emph{ACM Int. Conf.
  Multimedia}}. \bibinfo{pages}{2429--2437}.
\newblock


\bibitem[Zhang et~al\mbox{.}(2016)]%
        {zhang2016colorful}
\bibfield{author}{\bibinfo{person}{Richard Zhang}, \bibinfo{person}{Phillip
  Isola}, {and} \bibinfo{person}{Alexei~A Efros}.}
  \bibinfo{year}{2016}\natexlab{}.
\newblock \showarticletitle{Colorful image colorization}. In
  \bibinfo{booktitle}{\emph{Eur. Conf. Comput. Vis.}} Springer,
  \bibinfo{pages}{649--666}.
\newblock


\bibitem[Zhang et~al\mbox{.}(2018)]%
        {zhang2018unreasonable}
\bibfield{author}{\bibinfo{person}{Richard Zhang}, \bibinfo{person}{Phillip
  Isola}, \bibinfo{person}{Alexei~A Efros}, \bibinfo{person}{Eli Shechtman},
  {and} \bibinfo{person}{Oliver Wang}.} \bibinfo{year}{2018}\natexlab{}.
\newblock \showarticletitle{The unreasonable effectiveness of deep features as
  a perceptual metric}. In \bibinfo{booktitle}{\emph{IEEE Conf. Comput. Vis.
  Pattern Recog.}} \bibinfo{pages}{586--595}.
\newblock


\bibitem[Zhang et~al\mbox{.}(2017)]%
        {zhang2017real}
\bibfield{author}{\bibinfo{person}{Richard~Yi Zhang}, \bibinfo{person}{Jun~Yan
  Zhu}, \bibinfo{person}{Phillip Isola}, \bibinfo{person}{Xinyang Geng},
  \bibinfo{person}{Angela~S Lin}, \bibinfo{person}{Tianhe Yu}, {and}
  \bibinfo{person}{Alexei~A Efros}.} \bibinfo{year}{2017}\natexlab{}.
\newblock \showarticletitle{Real-time user-guided image colorization with
  learned deep priors}.
\newblock \bibinfo{journal}{\emph{ACM Trans. Graph.}} \bibinfo{volume}{36},
  \bibinfo{number}{4} (\bibinfo{year}{2017}), \bibinfo{pages}{119}.
\newblock


\bibitem[Zhang et~al\mbox{.}(2021b)]%
        {zhang2021beyond}
\bibfield{author}{\bibinfo{person}{Yonghua Zhang}, \bibinfo{person}{Xiaojie
  Guo}, \bibinfo{person}{Jiayi Ma}, \bibinfo{person}{Wei Liu}, {and}
  \bibinfo{person}{Jiawan Zhang}.} \bibinfo{year}{2021}\natexlab{b}.
\newblock \showarticletitle{Beyond brightening low-light images}.
\newblock \bibinfo{journal}{\emph{Int. J. Comput. Vis.}} \bibinfo{volume}{129},
  \bibinfo{number}{4} (\bibinfo{year}{2021}), \bibinfo{pages}{1013--1037}.
\newblock


\bibitem[Zhang et~al\mbox{.}(2019)]%
        {zhang2019kindling}
\bibfield{author}{\bibinfo{person}{Yonghua Zhang}, \bibinfo{person}{Jiawan
  Zhang}, {and} \bibinfo{person}{Xiaojie Guo}.}
  \bibinfo{year}{2019}\natexlab{}.
\newblock \showarticletitle{Kindling the darkness: A practical low-light image
  enhancer}. In \bibinfo{booktitle}{\emph{ACM Int. Conf. Multimedia}}.
  \bibinfo{pages}{1632--1640}.
\newblock


\bibitem[Zhang et~al\mbox{.}(2022)]%
        {zhang2022deep}
\bibfield{author}{\bibinfo{person}{Zhao Zhang}, \bibinfo{person}{Huan Zheng},
  \bibinfo{person}{Richang Hong}, \bibinfo{person}{Mingliang Xu},
  \bibinfo{person}{Shuicheng Yan}, {and} \bibinfo{person}{Meng Wang}.}
  \bibinfo{year}{2022}\natexlab{}.
\newblock \showarticletitle{Deep Color Consistent Network for Low-Light Image
  Enhancement}. In \bibinfo{booktitle}{\emph{IEEE Conf. Comput. Vis. Pattern
  Recog.}} \bibinfo{pages}{1899--1908}.
\newblock


\bibitem[Zhao et~al\mbox{.}(2020)]%
        {zhao2020pixelated}
\bibfield{author}{\bibinfo{person}{Jiaojiao Zhao}, \bibinfo{person}{Jungong
  Han}, \bibinfo{person}{Ling Shao}, {and} \bibinfo{person}{Cees~GM Snoek}.}
  \bibinfo{year}{2020}\natexlab{}.
\newblock \showarticletitle{Pixelated semantic colorization}.
\newblock \bibinfo{journal}{\emph{Int. J. Comput. Vis.}} \bibinfo{volume}{128},
  \bibinfo{number}{4} (\bibinfo{year}{2020}), \bibinfo{pages}{818--834}.
\newblock


\bibitem[Zhao et~al\mbox{.}(2022)]%
        {zhao2022fcl}
\bibfield{author}{\bibinfo{person}{Suiyi Zhao}, \bibinfo{person}{Zhao Zhang},
  \bibinfo{person}{Richang Hong}, \bibinfo{person}{Mingliang Xu},
  \bibinfo{person}{Yi Yang}, {and} \bibinfo{person}{Meng Wang}.}
  \bibinfo{year}{2022}\natexlab{}.
\newblock \showarticletitle{FCL-GAN: A Lightweight and Real-Time Baseline for
  Unsupervised Blind Image Deblurring}.
\newblock \bibinfo{journal}{\emph{arXiv preprint arXiv:2204.07820}}
  (\bibinfo{year}{2022}).
\newblock


\bibitem[Zhao et~al\mbox{.}(2021)]%
        {zhao2021unsupervised}
\bibfield{author}{\bibinfo{person}{Suiyi Zhao}, \bibinfo{person}{Zhao Zhang},
  \bibinfo{person}{Richang Hong}, \bibinfo{person}{Mingliang Xu},
  \bibinfo{person}{Haijun Zhang}, \bibinfo{person}{Meng Wang}, {and}
  \bibinfo{person}{Shuicheng Yan}.} \bibinfo{year}{2021}\natexlab{}.
\newblock \showarticletitle{Unsupervised color retention network and new
  quantization metric for blind motion deblurring}.
\newblock \bibinfo{journal}{\emph{TechRxiv Preprint}} (\bibinfo{year}{2021}),
  \bibinfo{pages}{1--20}.
\newblock


\bibitem[Zheng et~al\mbox{.}(2022)]%
        {zheng2022enhancement}
\bibfield{author}{\bibinfo{person}{Naishan Zheng}, \bibinfo{person}{Jie Huang},
  \bibinfo{person}{Qi Zhu}, \bibinfo{person}{Man Zhou}, \bibinfo{person}{Feng
  Zhao}, {and} \bibinfo{person}{Zheng-Jun Zha}.}
  \bibinfo{year}{2022}\natexlab{}.
\newblock \showarticletitle{Enhancement by Your Aesthetic: An Intelligible
  Unsupervised Personalized Enhancer for Low-Light Images}. In
  \bibinfo{booktitle}{\emph{ACM Int. Conf. Multimedia}}.
  \bibinfo{pages}{6521--6529}.
\newblock


\end{thebibliography}

\clearpage
\appendix
\begin{figure}
    \centering
    \includegraphics[width=1.\columnwidth]{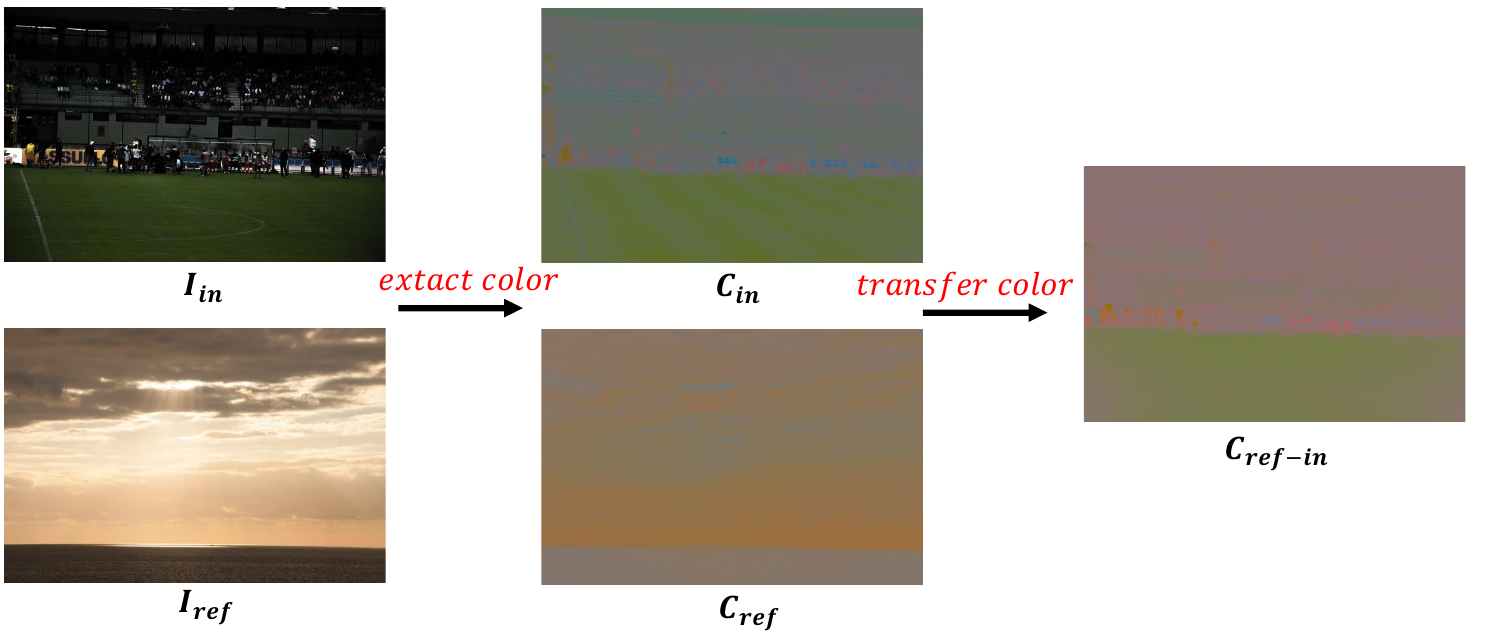}
     \vspace{-0.7cm}
    \caption{ The process of color adaptation.}
    \label{fig:color_transfer}
    \vspace{-0.4cm}
\end{figure}

\begin{figure}
    \centering
    \includegraphics[width=0.8\columnwidth]{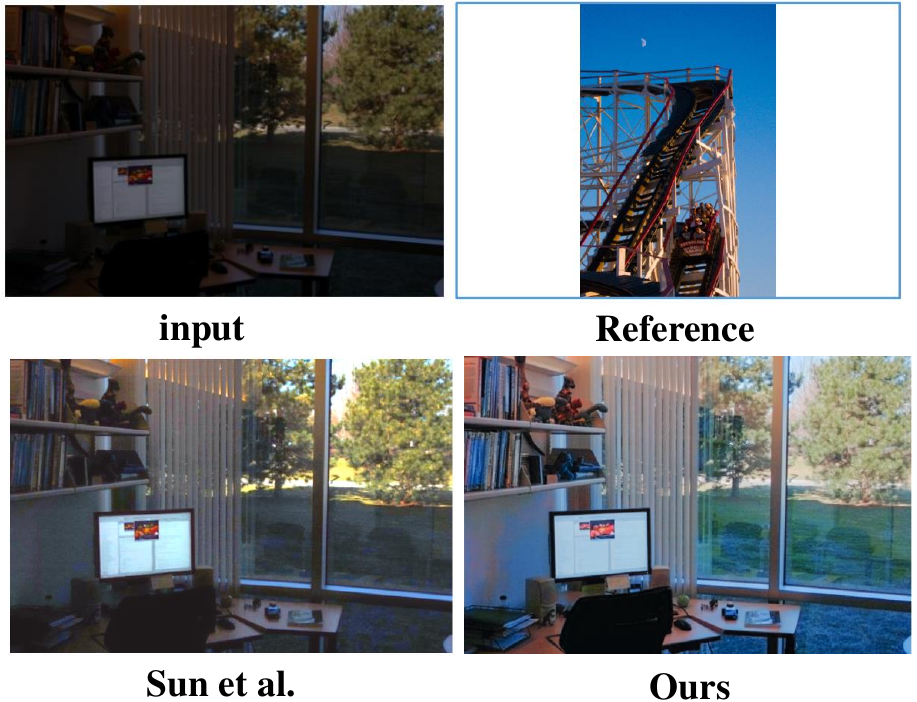}
     \vspace{-0.4cm}
    \caption{Visual comparison with other Su et al. \cite{sun2021enhance}.}
    \label{fig:com_cus}
    \vspace{-0.4cm}
\end{figure}

\section{Additional Ablation Studies}

\noindent\textbf{Loss functions.} We further conduct an ablation study on loss functions.
\begin{itemize} 
\item ``W/o $\mathcal{L}_{ssim}$" removes ssim loss in brightening sub-task.  
\item ``W/o $\mathcal{L}_{tv}$" removes smooth loss in brightening sub-task. 
\item ``W/o $\mathcal{L}_{per}$" removes perceptual loss in colorization sub-task. 
\end{itemize}
The results in Table \ref{tab:ablation_loss} demonstrate the effectiveness of adopted loss functions. 

\vspace{0.1cm}
\noindent\textbf{Color spaces.} In fact, there are several color spaces that can decompose an image into lightness and chrominance, such as HSV, HSI, Luv, Yuv, et al.. We conduct an ablation study on color spaces to verify the effectiveness of adopted Lab color space. However, since it is hard to quantize chrominance channels in other color spaces like \cite{zhang2016colorful} in Lab color space, we present the results of removing color classification loss in Lab color space for a fair comparison. As shown in Table \ref{tab:ablation_color_space}, the Lab achieves the best results. Actually, it is also the popular color space in the image colorization field.

% ablation losses
\begin{table}[]
\caption{Results of the loss functions ablation studies on LOL-real \cite{yang2021sparse} dataset. The best results are boldfaced and the second-best ones are underlined.}
    \label{tab:ablation_loss}
\renewcommand{\arraystretch}{1} % 可以调节, 1.2指高度是默认的1.2倍
    \centering
    
    \begin{tabular}{c c c c c}
         \hline
      Methods&  PSNR& SSIM& LPIPS & CSE (ratio) \\
     \hline
        w/o $\mathcal{L}_{ssim}$ & 22.40 & 0.8388 & 0.0801 & \underline{1.01} \\
      w/o $\mathcal{L}_{tv}$ & 22.02 & 0.8454  & 0.0739 & 1.39\\
      w/o $\mathcal{L}_{per}$ & \underline{22.92} &  \underline{0.8582 }& \underline{0.0729} & 1.99\\
     \hline
     \multicolumn{1}{c}{Ours} & \textbf{23.27} & \textbf{0.8637}  & \textbf{0.0566}& \textbf{1.00}\\
     \hline
    
    \end{tabular}
\end{table}

\begin{table}[]
\caption{Quantitative comparison with customized LLIE methods.}
    \label{tab:cmp_cus}
    \resizebox{\columnwidth}{!}{
\renewcommand{\arraystretch}{1} % 可以调节, 1.2指高度是默认的1.2倍
    \centering
    \begin{tabular}{c c c c c}
         \hline
      Methods&  PieNet \cite{kim2020pienet}& Sun te al. \cite{sun2021enhance}& TSFlow \cite{wang2022learning} & Ours \\
     \hline
       PSNR(dB) & 25.28 & 20.87 & 25.57 & 25.74 \\
     \hline
    \end{tabular}}
\end{table}

% ablation color space
\begin{table}[]
\caption{Results of the color space ablation studies on LOL-real \cite{yang2021sparse} dataset.  
The best results
are boldfaced and the second-best ones are underlined.}
    \label{tab:ablation_color_space}

\renewcommand{\arraystretch}{1} % 可以调节, 1.2指高度是默认的1.2倍
    \centering
    \begin{tabular}{c c c c c}
         \hline
      Methods&  PSNR& SSIM& LPIPS & CSE (ratio) \\
     \hline
        HSV & 22.02 & 0.8257 & 0.0809 & 1.12 \\
      HLS & 22.31 & \underline{0.8454}  & 0.0869 & 1.38\\
      Luv&22.02 &  0.8420 & \underline{0.0702} & \underline{1.06}\\
      Yuv& \underline{22.37} &  0.8416 & 0.0751 & 1.16\\
     \hline
     \multicolumn{1}{c}{Lab (ours)} & \textbf{23.21} & \textbf{0.8600}  & \textbf{0.0619}& \textbf{1.00}\\
     \hline
    \end{tabular}
\end{table}

\section{Comparison with other customized LLIE method}
We conduct the comparison with other customized LLIE methods in Table. \ref{tab:cmp_cus} and Fig. \ref{fig:com_cus}. Note that we only compare the visual result with Sun et al. \cite{sun2021enhance} since only their method is based on one reference image and opens the source code. As can be seen, the proposed method reaches accurate and flexible enhancement.

\section{Implementation of Color Adaptation }
The color adaptation is responsible for generating customized color guidance based on a reference image. We utilize a non-learning method \cite{reinhard2001color} to accomplish this process. Given an input low-light image $I_{in}$ and a reference image $I_{ref}$, we first transform them into CIELAB color space:
\begin{equation}
\begin{split}
    L_{in},a_{in},b_{in} = RGB2Lab(I_{in})\\
    L_{ref},a_{ref},b_{ref} = RGB2Lab(I_{ref})\\
\end{split}
\end{equation}
where $RGB2Lab(.)$ represents the color space transform function. Then, the mean value and standard deviation are used to transfer color:
\begin{equation}
\begin{split}
    a_{in} = a_{in}-mean(a_{in}) , b_{in} = b_{in}-mean(b_{in})\\
\end{split}
\end{equation}
\begin{equation}
\begin{split}
    a_{in} = a_{in}\times(std(a_{in})/std(a_{ref})) \\ 
    b_{in} = b_{in}\times(std(b_{in})/std(b_{ref})\\
\end{split}
\end{equation}
where $mean(.)$ and $std(.)$ represent to extract the mean value and standard deviation. Finally, the transferred color guidance $C_{ref-in} = cat(a_{in},b_{in})$. The visual result can be seen in Fig. \ref{fig:color_transfer}. Note that the focus of color adaptation is the chrominance information, we ignore the lightness components of two images.

\section{More Visual results}
\begin{figure*}
  \centering
  \captionsetup[subfloat]{labelformat=empty}
  \subfloat[ 
  ]{
    \includegraphics[width=1.\textwidth]{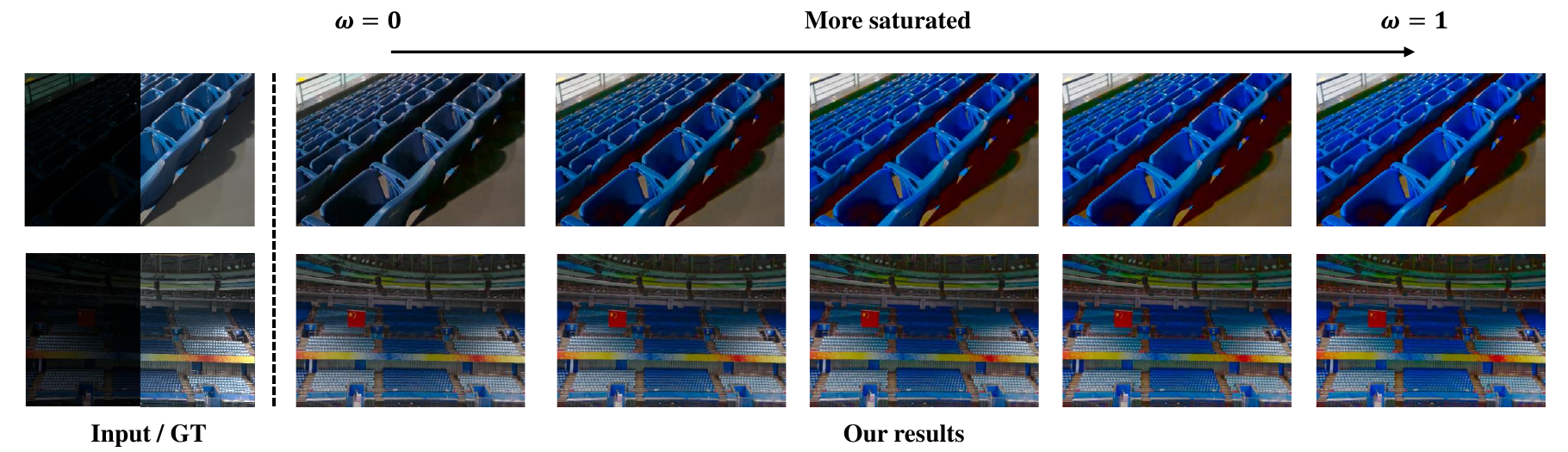}
  }
  \caption{Enhancement with diverse saturations. }
    \label{fig:diverse_saturations_supp}
\end{figure*}

\begin{figure*}
  \centering
  \captionsetup[subfloat]{labelformat=empty}
  \subfloat[ % (6.72/0.3461)
  ]{
    \includegraphics[width=1.\textwidth]{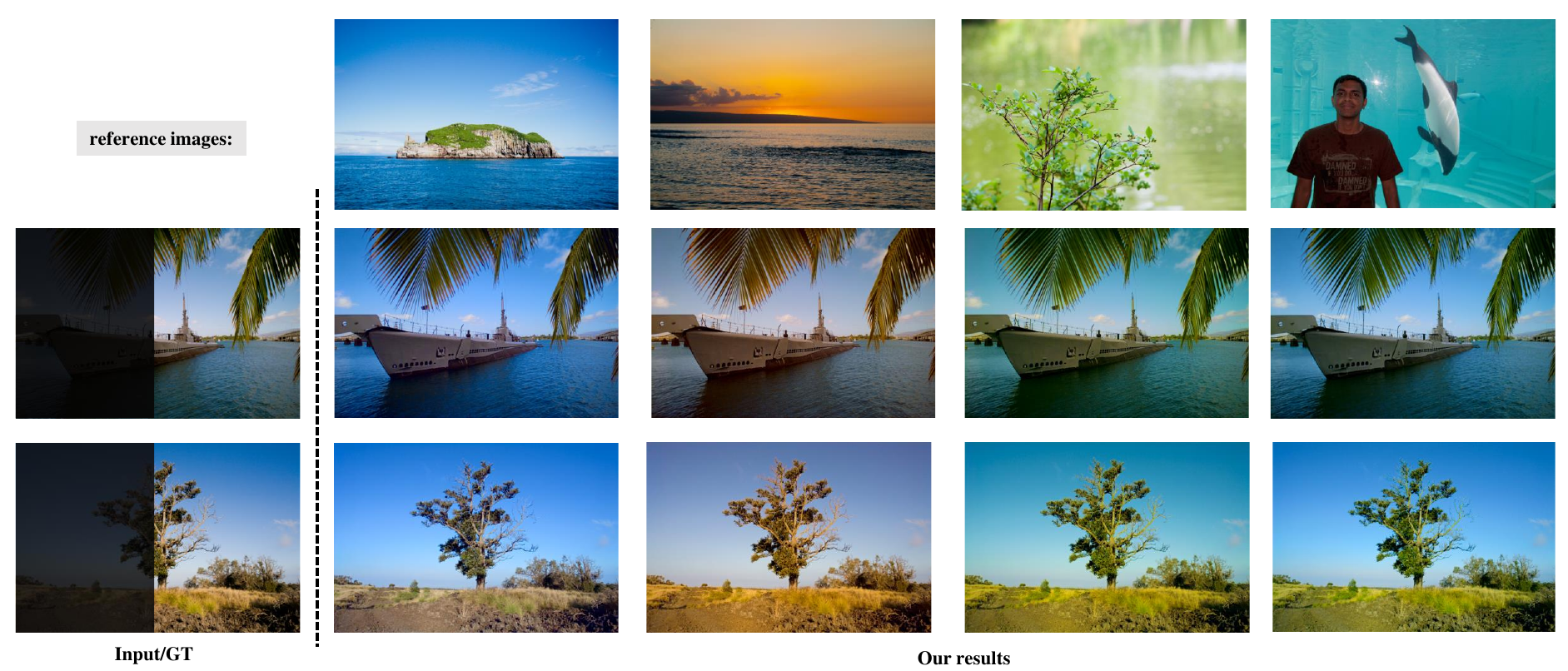}
  }
  \caption{Enhancement with diverse color styles. }
    \label{fig:diverse_styles_supp}
\end{figure*}

We present more customized enhancement results in Fig. \ref{fig:diverse_saturations_supp} (enhancement with diverse saturations) and Fig. \ref{fig:diverse_styles_supp} (enhancement with diverse color styles).


\end{document}